\pdfoutput=1
\documentclass[11pt]{article}
\usepackage{acl}

\usepackage{newtxtext}
\usepackage{newtxmath}
\usepackage{latexsym}
\usepackage[T1]{fontenc}
\usepackage[utf8]{inputenc}
\usepackage[nopatch=footnote]{microtype}
\usepackage{inconsolata}
\usepackage{graphicx}

\usepackage{hyperref}
\RequirePackage{doi}
\usepackage{booktabs}
\usepackage{tabularx}
\usepackage{enumitem}

\usepackage{linguex}
\usepackage{lscape}

\usepackage{caption}
\usepackage{subcaption}
\usepackage{comment}
\usepackage{multirow}
\usepackage{longtable}
\usepackage{rotating}
\usepackage{CJKutf8}
\usepackage{placeins}
\usepackage{colortbl}
\usepackage[absolute]{textpos}

\newcommand{\OS}{OpenSubtitles}
\newcommand{\atag}[1]{\begin{small}\textsf{#1}\end{small}}

\begin{document}

\title{Conversational Feedback in Scripted versus Spontaneous Dialogues:\\ A Comparative Analysis}

\newcommand*{\affaddr}[1]{#1} 
\newcommand*{\affmark}[1][*]{\textsuperscript{#1}}
\newcommand*{\email}[1]{\texttt{#1}}

\author{
 Ildikó Pilán\affmark[1], Laurent Prévot\affmark[2]\textsuperscript{,}\affmark[3], Hendrik Buschmeier\affmark[4], and Pierre Lison\affmark[1]\\[3mm]
\affaddr{\affmark[1] Norwegian Computing Center, Oslo, Norway}\\
\affaddr{\affmark[2] CEFC, CNRS, MEAE, Taipei, Taiwan}\\
\affaddr{\affmark[3] Aix Marseille Université \& CNRS, LPL, Aix-en-Provence, France}\\
\affaddr{\affmark[4] Faculty of Linguistics and Literary Studies, Bielefeld University, Bielefeld, Germany} \\[3mm]
\email{{\normalsize \{pilan,plison\}@nr.no \ \ \ \ \   laurent.prevot@univ-amu.fr \ \ \ \ \  hbuschme@uni-bielefeld.de}} 
} 

\maketitle

\begin{textblock*}{200mm}(5mm,5mm)
  {\color{gray}
  \noindent \textsf{In: \emph{Proceedings of the 25th Annual Meeting of the Special Interest Group on Discourse and Dialogue (SIGdial 2024)},\\ pp. 440–457, Kyoto, Japan. \url{https://doi.org/10.18653/v1/2024.sigdial-1.38}}}
\end{textblock*}

\begin{abstract}%
Scripted dialogues such as movie and TV subtitles constitute a widespread source of training data for conversational NLP models. However, there are notable linguistic differences between these dialogues and spontaneous interactions, especially regarding the occurrence of \textit{communicative feedback} such as backchannels, acknowledgments, or clarification requests. 
This paper presents a quantitative analysis of such feedback phenomena in both subtitles and spontaneous conversations. Based on conversational data spanning eight languages and multiple genres, we extract lexical statistics, classifications from a dialogue act tagger, expert annotations and labels derived from a fine-tuned Large Language Model (LLM). Our main empirical findings are that (1) communicative feedback is markedly less frequent in subtitles than in spontaneous dialogues and (2) subtitles contain a higher proportion of negative feedback. We also show that dialogues generated by standard LLMs lie much closer to scripted dialogues than spontaneous interactions in terms of communicative feedback. 
\end{abstract}

\section{Introduction}
\label{sec:introduction}

While the amount of text data available for training or fine-tuning LLMs is large and growing steadily, spoken conversational data remains relatively scarce. Although corpora of spontaneous spoken interactions have been collected for various languages \citep{Dingemanse2022}, those are generally of a modest size and limited to specific topics or tasks. Due to this scarcity of available data, a common approach for the development of conversational models is to rely on corpora of authored dialogues extracted from movie scripts \citep{danescu-niculescu-mizil-lee-2011-chameleons} or movie and TV subtitles \citep{Lison2018OpenSubtitles2018SR,davies2021tv}.

However, those dialogues are markedly different from spontaneous interactions. Most importantly, movie scripts and subtitles are explicitly written with the aim of \textit{narrating a story}. Subtitles must also abide to strict length constraints, and thus tend to only transcribe the most salient part of each turn. As a consequence, many conversational phenomena such as disfluencies \citep{shriberg1996disfluencies}, overlapping talk \citep{schegloff2000overlapping}, and backchannels \citep{Yngve1970} are either absent or uncommon in those dialogues, unless their presence happens to contribute to the storyline \citep{Berliner1999,ChepinchikjThompson2016}. 

This paper provides a quantitative analysis of how subtitles differ from spontaneous dialogues, focusing more specifically on \emph{conversational feedback} \citep{Allwood1992} and \emph{grounding} \citep{Clark1989} phenomena.
To highlight differences in linguistic properties between subtitles and spontaneous conversation corpora, we first compile a range of lexical statistics and use a dialogue act tagger to estimate the relative frequencies of various feedback signals. To obtain more fine-grained estimates on three core feedback categories, respectively \textit{Agreement / Acceptance}, \textit{Acknowledgement / Backchannel} and \textit{Negative Feedback}, we collect manual annotations on multiple dialogue samples and fine-tune a LLM on those annotations to automatically detect the presence of those feedback in our corpora. Finally, we apply the fine-tuned LLM on synthetic dialogues generated with standard autoregressive LLMs, and show that those dialogues are comparatively much closer to scripted dialogues than to spontaneous interactions when it comes to the frequency and type of conversational feedback. Those experiments are conducted for eight languages (English, Chinese, French, German, Hungarian, Italian, Japanese and Norwegian) for which corpora of spontaneous dialogues are readily available. 

The paper is structured as follows. Section~\ref{sec:related-work} reviews related work, and Section~\ref{sec:corpora} presents the corpora employed in our experiments. Section~\ref{sec:lexstats} describes the observed lexical distributions of feedback phenomena and Section~\ref{sec:tagging} compares them to estimates derived with a dialogue act tagger. In Section \ref{sec:annotations}, we describe the manual annotation of dialogue samples and the fine-tuning of an LLM to automate this process.  Finally, Section~\ref{sec:synthetic} describes the results of applying this LLM-based method to synthetic dialogues, and Section \ref{sec:conclusion} concludes. 

\section{Related Work}
\label{sec:related-work}

\subsection{Conversational Feedback and Grounding}

A key aspect of any communicative activity is the management of the common ground, a process often called \emph{conversational grounding} \citep{Clark1989}. The study of grounding and related phenomena, such as conversational feedback \citep{Allwood1992}, has been instrumental to cognitive approaches to communication \citep{Clark1996}, and to dialogue system development \citep{Traum1994,PaekHorvitz2000-TR,YaghoubzadehPitsch2015}. 

Feedback and grounding can happen at any of the \emph{levels of communication} that includes simple contact, perception, understanding and higher-level evaluation of what had been said \citep{Allwood1992,Clark1996}.  
Conversational feedback may appear at different positions in a dialogue. However, a number of corpus studies found that they have a tendency to occur at specific places, mostly where they cause little interference \citep{kjellmer2009we}. These places of occurrence have also been referred to as \emph{Feedback Relevant Spaces} \citep{HeldnerHjalmarsson2013,Howes.Eshghi_JLLI_2021}.
Although, arguably, any utterance relates directly or indirectly to grounding (through implicit and high level pragmatic inference, \citealt{Clark1989}), \emph{acknowledgments} and other positive feedback signals (see Ex.~\ref{ex:ack}), along with \emph{repair} (see Ex.~\ref{ex:map}), have been identified as the most prominent grounding mechanisms \citep{Jefferson1972,Bunt1994}. Their frequency in human-human dialogue is known to be very high \citep[e.g.,][]{StolckeRies2000} and universal across languages \citep{LiesenfeldDingemanse2022,DingemanseRoberts2015}. These conversational signals, while they do not cover all grounding phenomena, can therefore be seen as a useful proxy to quantify feedback in a dialogue. 

    \ex. \label{ex:ack} 
    \textbf{A:} and uh it really does irk me to see those guys out there uh you know making that ///much money///\\
    \textbf{B:} ///yeah///\footnote{Notation: ///text/// produced in overlap with the speech of the other speaker. From Switchboard \citep{godfrey1992switchboard}}

Recent works have emphasized the role of feedback and grounding signals in their study of human-human conversations \citep{Fusaroli2017,Dideriksen2022,Dingemanse2022} as well as human-agent interaction \citep{VisserTraum2014,HoughSchlangen2016,BuschmeierKopp2018.AAMAS,AxelssonBuschmeier2022}. 

The literature tends to merge the two closely related concepts of \emph{backchannels} and \emph{acknowledgments}. Backchannels \citep{Yngve1970}, or \emph{continuers} \citep{schegloff1982discourse}, are not positioned on the main channel, but uttered by the ``listener'', often as low intensity unobtrusive overlapping speech \citep{HeldnerEdlund2010} or non-verbally \citep{AllwoodKopp2007,TruongPoppe2011}. Acknowledgments, on the other hand, have a slightly broader, functional definition of minimal positive feedback \citep{Jefferson1984,Allwood1992}. 

There is a large body of work on lexical markers, also called \emph{cue phrases} or \emph{discourse markers} \citep{Jefferson1984,Allwood1992,Muller2003}, since they present interesting linguistic features and constitute convenient explicit cues for detecting feedback utterances automatically \citep{Jurafsky1998,Gravano2011,Prevot2015}. \citet{Gravano2011} developed a list of affirmative cue words made of \emph{alright, mm-hm, okay, right, uh-huh, yeah}. Form-Function studies of similar lists have been made at least for Swedish \citep{Allwood1988}, U.S.~English \citep{Ward2006}, and French \citep{Prevot2015}.

Few studies have, however, concentrated on direct negative feedback associated with rejection and corrective dialogue acts. Although \citet{Allwood1992} suggests a polarity dimension for characterizing feedback, most recent studies have focused on positive feedback. Indeed, in collaborative dialogue and everyday conversations, which are the two genres dominating available datasets, positive feedback constitutes the large majority of explicit feedback \citep[e.g.,][]{Malisz-etal-2016}. 
Negative feedback is instead often expressed constructively, using repair mechanisms, specifically \emph{clarification requests} \citep{Purver2004}. 
These may rely on simple lexical cues (e.g., for English, \emph{pardon?, huh?}), sluices (such as \emph{what?, who?}), or on clarification ellipsis, as in the following example \citep{Fernandez2007}:

    \ex. \label{ex:map}
    \textbf{A:} and then we're going to turn east\\
    \textbf{B:} mmhmm\\
    \textbf{A:} not straight east slightly sort of northeast\\
    \textbf{B:} slightly northeast?\footnote{From HCRC Map Task \citep{anderson1991hcrc}.}

The occurrence of feedback signals in dialogue transcriptions can be detected using various types of sequence labeling models from classical hidden Markov models \citep{stolcke-etal-2000-dialogue} to modern neural architectures and large language models \citep{liu-etal-2017-using-context,noble-maraev-2021-large}.

\subsection{Analysis of Subtitles}

Subtitles are typically short written text snippets and they accompany audiovisual content on the screen. They are often subject to condensation and normalization, where non-standard verbal elements (repetitions, signs of hesitation etc.) are omitted or replaced by more standard alternatives \citep{gottlieb2012subtitles} due to constraints on the length, readability and writing conventions. As subtitles are displayed alongside audiovisual content, viewers can typically recover omitted dialogue-relevant cues from the accompanying images and sounds.  
\emph{Interlingual subtitling} -- where the original language of the audio is different from the subtitling language -- differs somewhat from \emph{intralingual subtitling}, which is meant for same-language audio and subtitles which also records non-verbal elements writing for the benefit of hearing impaired audiences or non-native speakers \citep{gottlieb2012subtitles}. 

\citet{Ruehlemann2020} compared real conversations and scripted ones and observed that continuers were absent from the latter. \citet{Prevot2019} compared data from the \emph{Open Subtitles} corpus \citep{Lison2016,Lison2018OpenSubtitles2018SR} in English, French and Mandarin with both written and conversational corpora and found that \OS{} occupied an intermediate position between written and conversational data in terms of lexical and syntactic features. This paper builds upon those earlier works but focuses specifically on communicative feedback, using a combination of lexical statistics, manual and automate annotations to quantify its frequency in various corpora. 

\section{Corpora}
\label{sec:corpora}

We rely on data from both \OS{} and existing, publicly available corpora of real conversations covering eight different languages (see Table \ref{tbl:spontaneous_data}). 

\subsection{Spontaneous Dialogues}

\paragraph{German (de)} We use the Hamburg MapTask corpus \citep{HZSK2010}, in which twelve dyads of (L2) speakers of German engage in dyadic task-oriented short dialogues. 

\paragraph{English (en)} For English, we use 
    Switchboard (SWBD), consisting of dyadic topic oriented phone conversation \citep{godfrey1992switchboard} as well as Fisher \citep{Cieri2004} for some experiments; 
    AMI, with multi-party multimodal task-oriented dialogues \citep{Carletta2007};
    HCRC MapTask (MT) comprising dyadic task-oriented short dialogues \citep{anderson1991hcrc}; and 
    STAC, a multi-party negotiation chat corpus \citep{Asher2016}.

\paragraph{French (fr)} We include 
    CID, consisting of dyadic, 1-hour long, loosely topic-oriented face-to-face conversations with 16 participants \citep{Blache2017};
    French MapTask with 16 participants \citep{Gorisch2014}; and
    Aix-DVD, dyadic face-to-face conversations about movie preferences of 16 participants \citep{Prevot2016}.

\paragraph{Hungarian (hu)} We employ BUSZI-2 corpus \citep[Budapest Sociolinguistic Interview,][]{varadi_2003}, where 50 participants with different educational levels participated in a 30-minute directed conversation and then performed language tasks (e.g. grammaticality judgments). 

\paragraph{Italian (it)} We use the CLIPS corpus \citep{savy2009diatopic}, consisting of both a map task and a difference spotting task between images. We exclude dialogues with a high proportion ($>10\%$) of utterances with dialectal words.

\paragraph{Japanese (ja)} This language is represented by the transcripts of the CallHome Japanese corpus \citep{den2000callhome} consisting of 120 unscripted telephone conversations between native speakers, mostly family members or close friends.

\paragraph{Norwegian (no)}
We use the \href{http://www.tekstlab.uio.no/nota/oslo/english.html}{NoTa-Oslo} corpus \citep{johannessen-etal-2007-advanced}, containing interviews and conversations from 2004--2006 with 166 informants from the Oslo area. The dialogues consist of 10-minute semi-formal interviews and 30-min informal dialogues with other informants.

\paragraph{Mandarin Chinese (zh)}
The source of our Mandarin Chinese data was CALLHOME \citep{wheatley1996callhome} consisting of unscripted telephone conversations between native speakers. 

\begin{table*}[t!] 
\small
\begin{tabularx}{\textwidth}{lXXXXXXXXX}
	\toprule  
	   \textbf{Language}  & de & en    & fr  & hu  & it  & ja  & no  & zh  & \textbf{total} \\
	\midrule
\textbf{\# Spontaneous dialogues} & 24 & 2766  & 48  & 50  & 88  & 120 & 259 & 120 & \textbf{3475} \\ 
\textbf{\# Utterances}    & 4K & 373K  & 27K & 31K & 24K & 39K & 86K & 18K & \textbf{602K}  \\ 
\midrule
    \textbf{\# Subtitles}     & 98   & 100  & 100  & 68  & 95   & 74   & 87  & 93 & \textbf{715}  \\
    \textbf{\# Utterances} & 131K & 140K & 126K & 93K & 138K & 106K & 98K & 114K & \textbf{946K} \\
	\bottomrule
\end{tabularx}
\caption{Overview of dialogue data sources for both spontaneous conversations and subtitles employed in this paper.}
\label{tbl:spontaneous_data} 
\end{table*}

\subsection{Subtitles}
\label{ssec:subtitles}

The scripted dialogues are extracted from \href{https://opus.nlpl.eu/OpenSubtitles-v2018.php}{OpenSubtitles 2018} \citep{Lison2018OpenSubtitles2018SR}, a large collection of over 3.7 million subtitles (amounting to $\approx$~22.1 billion words) extracted from the OpenSubtitles.org database and covering 60 languages.
We include both (1) subtitles for the hearing impaired, where the subtitle language and the original audio language are identical and (2) subtitles for foreign audiences. 
The subtitles are then filtered according to several criteria. Only recent movies (year $\geq 1990$) are included to reflect contemporary language use, as is the case for the corpora of spontaneous conversations. We also omit subtitles with less then 100 utterances and exclude genres that are less relevant for this study (Documentary, Reality-TV, Biography, Sport, Musical, Music, Adult, Animation, Short and Game-Show).

We sample up to ten movies per audience type (hearing impaired vs. foreign audience) from the five largest genres, namely drama, comedy, crime, action, and romance.
Table \ref{tbl:spontaneous_data} shows the number of movies and utterances per language for the selected subtitles. Note that subtitles are typically segmented by dialogue turns or sentences instead of utterances. The term ``utterance'' should therefore be understood broadly in this paper.

This paper focuses on the textual aspects of grounding phenomena. While speech and non-linguistic aspects of communicative feedback (such as timing, intonation, gestures or gaze) are both important and well-studied, in particular for acknowledgements and backchannels, those information are not available in subtitles corpora, which are intrinsically limited to text transcriptions.

\section{Lexical Analysis}
\label{sec:lexstats}

Lexical statistics of acknowledgment cues gives us a first picture of the feedback frequency. Acknowledgments tend to be produced by the addressee (not the main speaker) and are therefore often short productions uttered in overlap and potentially with a lower voice. Out of those three properties (brevity, overlap, lower volume), only the first is practically measurable in our experiments, as the subtitles
are by construction text-based. 

\begin{figure*}[t!]
     \begin{subfigure}[b]{0.49\linewidth}
         \includegraphics[trim=6mm 2mm 0 7mm,clip, scale=0.45]{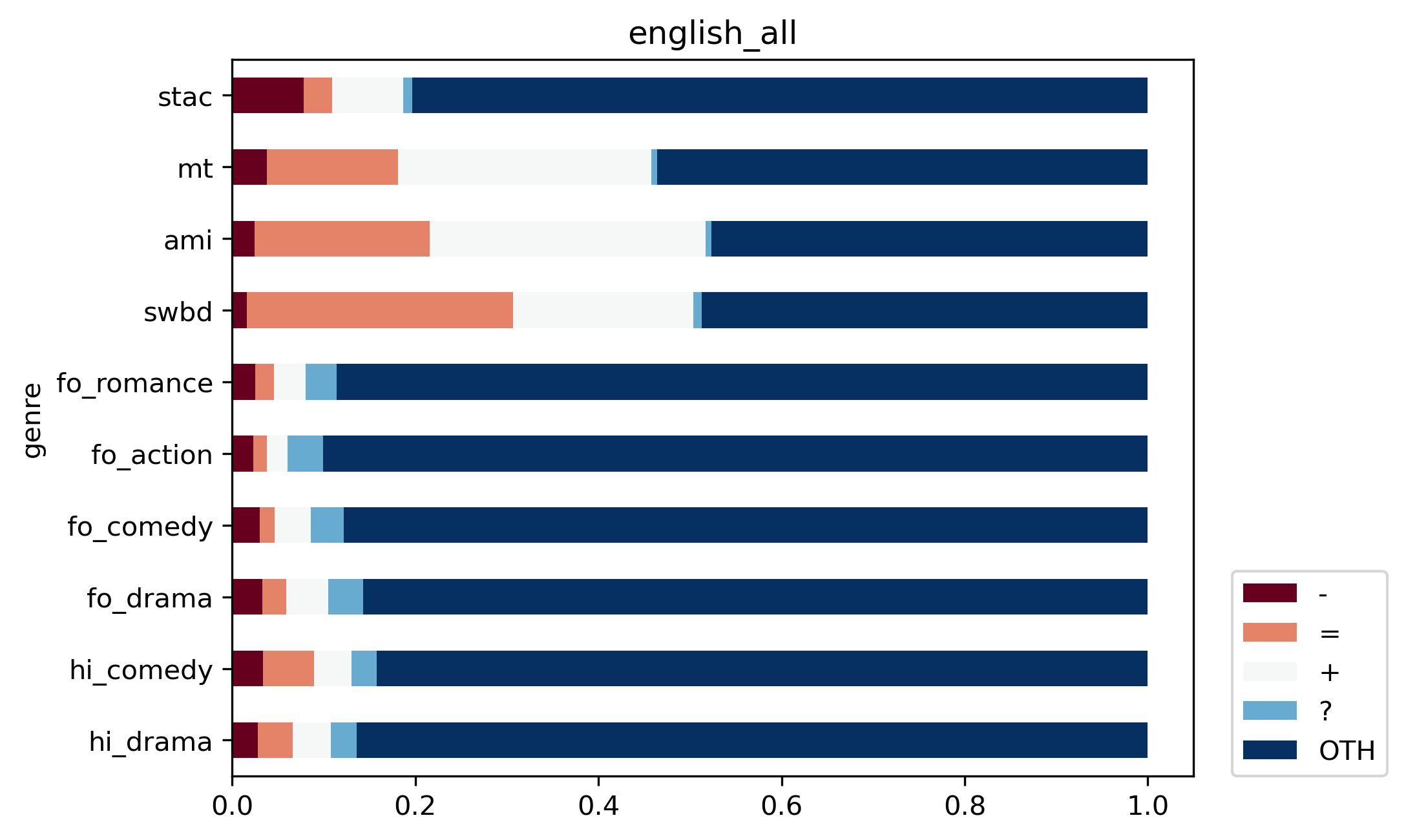}
         \caption{Absolute frequency}
         \label{fig:en_utts}
     \end{subfigure}
     \ \
     \begin{subfigure}[b]{0.49\linewidth}
         \includegraphics[trim=6mm 2mm 0mm 7mm, clip, scale=0.45]{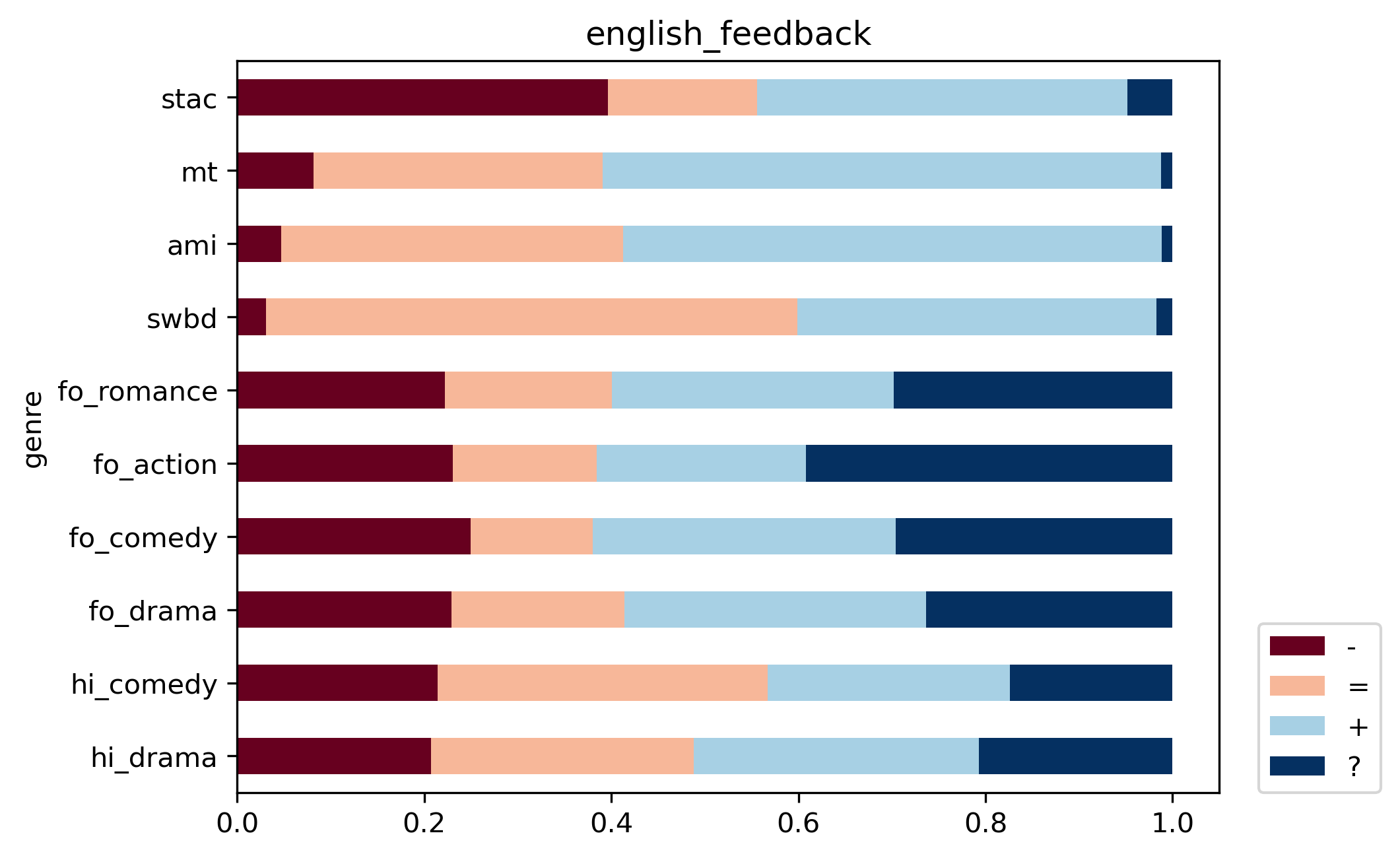}
         \caption{Relative frequency}
         \label{fig:en_feed}
     \end{subfigure}
        \caption{Frequency of conversational feedback of various types among utterances in the English corpora (both spontaneous and subtitles) based on manually curated lists of cue words to detect. Fig. (a) shows the absolute frequency while Fig. (b) zooms in on utterances labelled with at least one feedback. \textbf{+} denotes positive feedback/acknowledgement, \textbf{=} neutral/continuer feedback, \textbf{--} negative feedback, \textbf{?} clarification requests and ‘OTH’ is for other utterances. \texttt{fo} and \texttt{hi} respectively stand for ‘foreign audience’ and ‘hearing-impaired’ subtitles. Corpora without these prefixes are spontaneous dialogues.}
        \label{fig:lextstats_en}
\end{figure*}

Given their relation to acknowledgments, we first analyse ``very short utterances'' \citep{EdlundHeldner2009}, defined here as three tokens or less.
Feedback is also very well represented at initial positions of longer turns/contributions. 
We therefore targeted two locations: {\it very short utterances} (all tokens) and {\it initial positions} (one token) of all other utterances. Comparing term frequencies between these locations and the overall corpus allowed us to compile language-specific lists of \emph{cue words}. 
Those lists of cue words (presented in Table \ref{tab:cue_words} in the Appendix) are divided into four core classes of feedback:
\begin{itemize}[noitemsep]
    \item positive feedback/acknowledgment (\textbf{+})
    \item neutral/continuer (\textbf{=})
    \item negative feedback (\textbf{--})
    \item clarification request (\textbf{?}).
\end{itemize}

\begin{figure}[t!]
\vspace{1mm} 
         \includegraphics[trim=6mm 2mm 0 7mm,clip, scale=0.43]{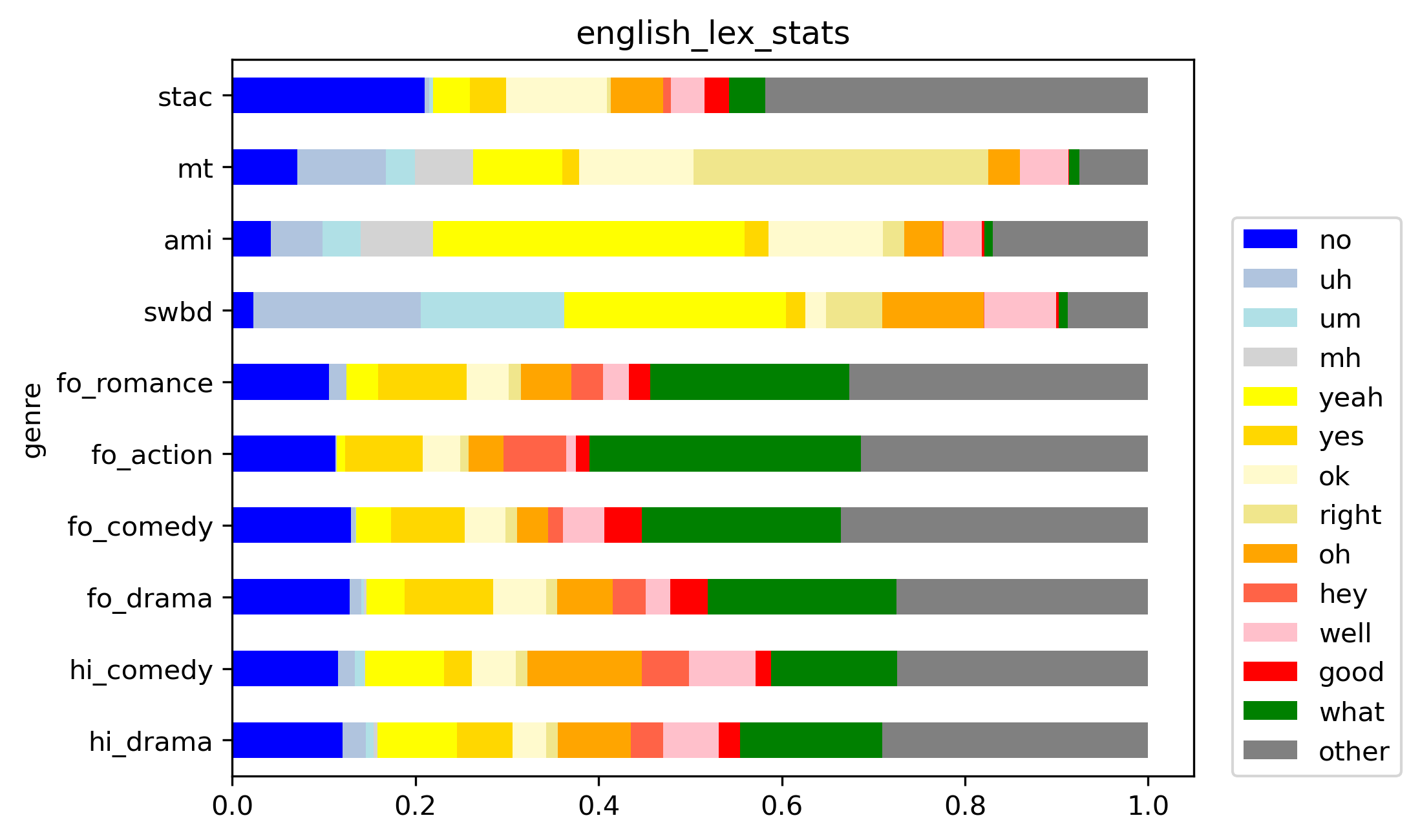}
        \caption{Most common lexical items associated with communicative feedback, as detected through manually curated lists of cue words in English, factored by corpus.\vspace{-3mm}}
        \label{fig:lexical_items}
\end{figure}

We plot in Figure~\ref{fig:lextstats_en} the frequencies of those feedback classes in each corpus, either in terms of absolute frequency (left side) or by looking at the relative proportions of the feedback classes (right side). Figure~\ref{fig:lexical_items} shows the lexical distribution of the most frequent lexical items observed in the utterances of plot (b) for English.

We observe that the statistics based on cue words differ substantially between subtitles and spontaneous dialogues. This difference is observed across all languages and sub-genres, (see Appendix~\ref{sec:appendix} for other languages). We sought to identify and reduce other sources of variation between corpora. STAC, as a chat corpus, exhibits different patterns than other dialogue corpora, notably due to the presence of emojis. Similarly, for English and French, we explored the impact of politeness expression (highly frequent in \OS{}). Those peculiarities did not, however, change the overall picture of our analysis (see Figure~\ref{fig:politeness} in Appendix~\ref{sec:appendix}).

One key difference between  real dialogues and subtitles relates to the overall frequency of feedback cues, which is much higher in spontaneous dialogues (40--50\%) than in subtitles (10--20\%), as observed in Figure~\ref{fig:lextstats_en}(a). Furthermore, as shown in Figure~\ref{fig:lextstats_en}(b), feedback in spontaneous dialogues consists mostly in positive or neutral ({\it continuers}) feedback, while subtitles have few neutral signals but seem to exhibit a much higher proportion of negative feedback and clarification requests.

We compared our English cue word lists against the annotations in Switchboard. After grouping feedback-related labels into a single \emph{Feedback} category,
we find that the cue word lists yield an $F_1$ score of 0.76.

\section{Dialogue Act Tagging}
\label{sec:tagging}

Although lexical statistics do highlight substantial differences in subtitles and spontaneous dialogues, they remain imprecise estimates, as many cue words related to feedback tend to be ambiguous. In this section, we refine our analysis using a dialogue act tagging model trained on the  DAMSL-Switchboard corpus. 

\subsection{Data}

We map the original set of Switchboard (SWBD) tags, and their clustered DAMSL-SWBD equivalents, into five coarse dialogue act (DA) classes: \emph{Forward looking}, \emph{Yes/no answers}, \emph{Assessment}, \emph{Backchannel} and \emph{Other}. The two classes most directly relevant for feedback, namely \emph{Backchannel} and \emph{Assessment}, are inspired, in part, by \citet{mezza-etal-2018-iso}. 
Distinguishing between these two feedback-related classes is also motivated by \citet{goodwin1986between}, who outline a number of positional and functional differences between these.
The \emph{Backchannel} category consists of the SWBD-DAMSL labels\footnote{\href{https://web.stanford.edu/~jurafsky/ws97/manual.august1.html}{web.stanford.edu/~jurafsky/ws97/manual.august1.html}} \emph{Acknowledge (Backchannel)}, (SWBD tag \atag{b}), \emph{Backchannel in question form}  (\atag{bh}), \emph{Response Acknowledgment} (\atag{bk}), \emph{Summarize/reformulate} (\atag{bf}) and \emph{Signal-non-understanding} (\atag{br}). As this latter tag suggests, negative feedback signals are also part of the \emph{Backchannel} category, since they are too few to reliably learn a separate class from.
The \emph{Assessment} category comprises not only the labels \emph{Agree/Accept} (\atag{aa}), but also \emph{Appreciation} (\atag{ba}) and \emph{Exclamation} (\atag{fe}).
The forward looking category contains utterances expressing explanations, instructions and suggestions as well as questions.
Table~\ref{tbl:swbd_mapped} in Appendix \ref{sec:appendix_tagger} shows the distribution of instances per label and their SWBD tag.

\begin{table*}[t!]
\begin{center} 
\begin{tabularx}{\textwidth}{Xllllllllr}  
    \toprule
    \bf DA group & \bf Data & \bf de & \bf en & \bf fr & \bf hu & \bf it & \bf ja & \bf no & \bf zh \\
    \midrule   
    \bf Assessment   & \bf \textsc{SpConv}   & 16.50 & 9.11  & 4.62  & 15.49 & 12.64 & 15.74 & 17.05 & 6.96 \\
    & \bf \textsc{Subs}     & 9.08  & 7.07 & 7.72 & 9.29 & 8.34 & 6.48 & 6.53 & 5.00 \\
    \bf Backchannel  &  \bf \textsc{SpConv}  & 11.57 & 10.79 & 11.96 & 4.28  & 5.73  & 18.96 & 2.67  & 5.65 \\
    & \bf \textsc{Subs}  & 3.49  & 3.72 & 3.44 & 3.48 & 3.45 & 3.74 & 3.47 & 3.00 \\
    \bf Yes/no answer & \bf \textsc{SpConv}  & 2.22  & 1.15  & 1.24  & 4.00  & 6.55  & 2.84  & 5.09  & 1.00 \\
     & \bf \textsc{Subs}  & 1.97  & 1.37 & 1.68 & 1.47 & 1.38 & 1.15 & 2.32 & 0.76 \\
    \bottomrule
\end{tabularx}
\caption{Proportions (\%) of the relevant dialogue act groups detected by the BERT-based dialogue act tagger in the spontaneous conversation (\textsc{SpConv}) and in the subtitle (\textsc{Subs}) corpora.}
\label{tbl:inference_res_conv}
\end{center}
\end{table*}

\subsection{Model Training}

We fine-tune the monolingual {\small \texttt{bert-base-cased}} pre-trained model \citep{Devlin2019BERTPO} using 80\% of the Switchboard data as training and 20\% for development and testing. We set up the task as a sequence classification problem, including the preceding utterance as context. 
We train the model with a batch size of $8$, a learning rate of $4E\!-\!5$ and default values for the other parameters. 
We run and compare three different random seeds, yielding similar performance. 
To improve recall, we also adjust the probability thresholds for the feedback classes.

The model performs relatively well on the Switchboard test set, yielding an accuracy of $0.81$. The  $F_1$ scores for the \textit{Assessment} and \textit{Backchannel} classes are respectively 0.59 and 0.83. This score difference may be due to \textit{Backchannel} instances being better represented in the training data, as well as some label confusion between the \textit{Assessment} and the \textit{Yes/No question} categories.

\subsection{Empirical Results}

We then use the trained dialogue act tagger to detect conversational feedback signals in both the spontaneous dialogue and subtitles.
For non-English corpora, we machine translate the data using the Google Translate API.
Feedback-annotated conversational corpora is non-existent for most languages and the quality of current MT systems is generally considered high enough to serve as a viable alternative \citep{isbister-etal-2021-stop}.
 
Table~\ref{tbl:inference_res_conv} presents the empirical results obtained with our dialogue act tagger on both spontaneous dialogues and subtitle corpora. We observe that backchannels are considerably more frequent (by a factor three) in spontaneous dialogues than in subtitles for half of the languages -- which is in line with the results of our lexical analysis in Section~\ref{sec:lexstats}. 
The number of utterances labeled as \emph{Assessment} differs less, but subtitles still seem to contain less of this feedback type in almost all genres and languages except French (see Appendix~\ref{sec:appendix_tagger} for details). Given that the tagger is only trained on a single corpus, some of the differences found may also be attributed to the generalization ability of the tagger to certain domains. We therefore also conduct some manual error analysis.

\subsection{Error Analysis}

In general, the proportion of the \emph{Backchannel} category for the spontaneous conversations is lower for Hungarian, Italian, Norwegian and Mandarin than for the other languages. This is likely due to the use of infrequent spelling variants of backchannel signals such as \emph{hmm, mh}. 
We have also found that the tagger has difficulties detecting feedback when they are part of longer utterances, whether they appear in an utterance-initial position or not. We also observe a general tendency to associate sentence-final question marks to feedback cues. When inspecting the most frequent utterances tagged as feedback, we also notice that short utterances pose some challenges for machine translation due to polysemy, e.g., Cosa? ``Thing?'', also translatable as ``What?'', in Italian.

\section{Further Annotations}
\label{sec:annotations}

The results from the dialogue tagger do show some clear trends regarding the extent to which communicative feedback is expressed in subtitles compared to spontaneous interactions. However, the use of DAMSL-Switchboard as sole source of training data is a limiting factor in our analysis, in particular when it comes to non-English dialogues, which must be machine-translated prior to labeling. Furthermore, the tagger does not provide information about the frequency of negative feedback, although the lexical analysis from Section \ref{sec:lexstats} does seem to point towards a higher frequency of those communicative signals in subtitles. 

We therefore complement the analyses of the two previous sections with a manual annotation effort. To this end, we sample from each corpus a set of 300 utterances to annotate. However, as evidenced by the results of the previous sections, many utterances of our corpora do not seem to contain any communicative feedback. To ensure the annotation process can cover a sufficiently broad variety of feedback signals despite this class imbalance, we do not select the utterances purely at random, but select half among those marked as feedback-relevant by the cue words of Section \ref{sec:lexstats}, and the other half among those that do not.

\subsubsection*{Annotation Process}

We recruited 6 annotators with prior expertise in linguistic annotation and proficient in the language corresponding to the corpus to annotate. Those annotators were provided each utterance in its context, and were tasked to decide whether the utterance in question contains one of the following three categories of communicative feedback: defined in the annotation guidelines as such: 
\begin{description}[itemsep=0.1em]

    \item[AGREE\_ACCEPT]: indicates that the speaker agrees or accepts what has been said.
    
    \item[ACK\_BACK]:  indicates that the speaker is listening to her interlocutor, or at least heard what has been said, without necessarily agreeing with it or committing to its content. 
    
    \item[NEGATIVE\_FEEDBACK]: indicates that the speaker could not hear or understand her interlocutor, or even rejects or disagrees with what the other person has said. 

\end{description}

Answers to explicit questions should not be considered as feedback. Each utterance can be tagged with zero, one, or multiple feedback labels. These categories specifically target and distinguish between different conversational feedback phenomena and are therefore somewhat more comprehensive than the categories employed by the tagger of the previous section. There, similar categories were derived by merging the available feedback-relevant dialogue act labels from the SWBD annotations.   

A total of 24 corpus samples, each comprising 300 utterances, were annotated\footnote{The full set of annotated dialogue samples is available at \url{https://github.com/NorskRegnesentral/conv_feedback}.}. 
Three corpus samples (respectively for English, French and Chinese) were doubly annotated, and the Kappa's score of their agreement was found to be 0.59 for \textit{AGREE\_ACCEPT}, 0.42 for \textit{ACK\_BACK} and 0.54 for \textit{NEGATIVE\_FEEDBACK} across the 3 samples. This relatively low inter-annotator agreement illustrates the challenging nature of the annotation task, in particular due to the lack of explicit turn boundaries in subtitles, making it at times difficult to determine the context behind each utterance.

\subsection{Annotation Results}

\begin{figure}[t]
\vspace{-5mm}\hspace{-3mm}\includegraphics[scale=0.53]{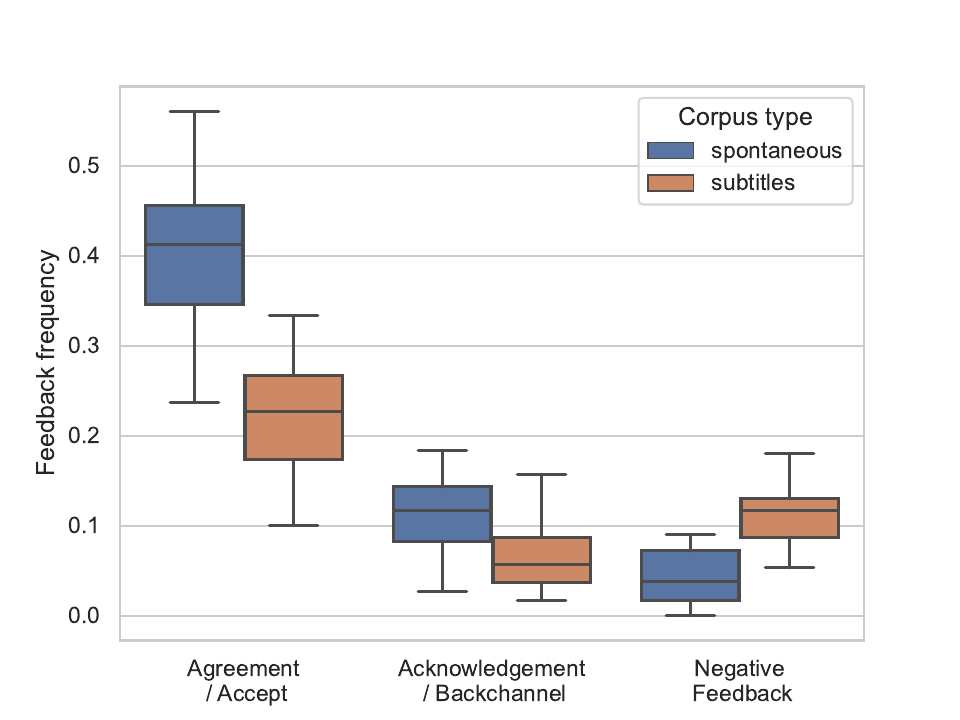}
\caption{Frequency of communicative feedback depending on the source of the dialogue sample (spontaneous interactions or  subtitles) and the category of feedback, based on annotations from human experts.\vspace{-3mm}}
\label{fig:frequency_in_sample}
\end{figure}

Figure \ref{fig:frequency_in_sample} illustrates the frequencies of the three feedback categories across the 24 annotated  samples. We observe again a lower proportion of both \textit{Agree / Accept} and \textit{Acknowledgement / Backchannel} feedbacks in the subtitles compared to real interactions. The proportion of \textit{Negative feedback} is, however, higher for the subtitles. We hypothesise that this may stem from the fact that disagreements between interlocutors are more interesting from the storytelling perspective, and are therefore more common in subtitles than in real interactions. 

We investigated whether subtitles for foreign audiences differed from subtitles written for the hearing impaired (as those often need to adhere more closely to the original on-screen conversation), but did not find any substantial disparity. 

\subsection{LLM-based Annotation}

The frequencies of Figure \ref{fig:frequency_in_sample} are obtained using the manually annotated dialogue samples. However, those samples only cover a small fraction of available corpora. Furthermore, as the sampling procedure relied on the use of cue-words to cover a sufficiently broad set of feedback types (see above), it is likely to overestimate the proportion of communicative feedback. To mitigate this bias, we fine-tune an instruction-tuned Gemma 2 model \citep{Gemma_Team2024-rz} to predict the probability of an utterance including one of the three defined feedback categories. The fine-tuning relied on LoRA \citep{Hu2021-kf} and included as instructions the annotation guidelines also provided to the human experts. The full set of 24 dialogue samples was used for the fine-tuning, each utterance being provided in its local dialogue context. For non-English utterances, we also include in the prompt an English translation of the utterance and its context, obtained using Google Translate.

\begin{figure}[t!]
\vspace{-5mm}\hspace{-2mm}\includegraphics[scale=0.53]{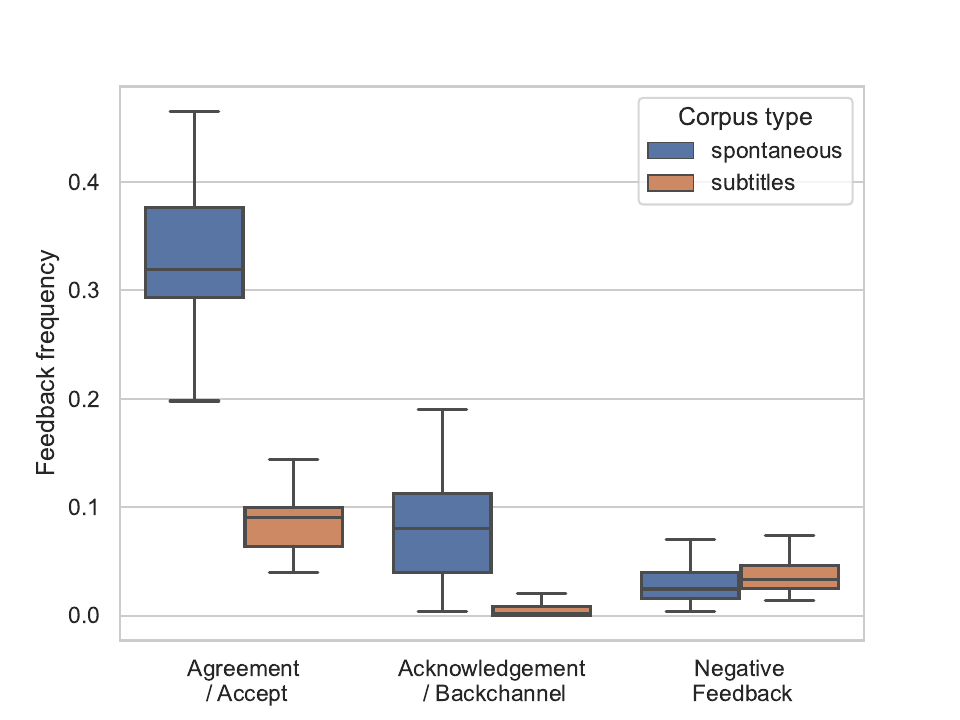}
\caption{Frequency of communicative feedback depending on the corpus type and category of feedback, based on the predictions of the fine-tuned Gemma 2 model trained on human annotations.}
\label{fig:frequency_in_corpus}
\end{figure}

The fine-tuned Gemma2 LLM was then applied to all corpora to predict whether their utterances contained one of the three categories of feedback defined above. The results are shown in Figure \ref{fig:frequency_in_corpus}. The proportions of communicative feedback are somewhat lower in the actual corpora than in the annotated samples (which is expected given how the dialogue samples were derived), but the overall trends remain similar to Figure \ref{fig:frequency_in_sample}. 

\section{Conversational Feedback in Synthetic Dialogues}
\label{sec:synthetic}

We conclude by investigating the occurrence of communicative feedback in synthetic dialogues generated with autoregressive language models. More precisely, we wish to analyze whether the communicative feedback generated by those models are closer to the patterns found in real interactions or to scripted dialogues such as subtitles. 

To this end, we use available GPT-2 models \citep{radford2019language} for the eight covered languages \footnote{The following pre-trained models are employed: \texttt{gpt2-base} (English), \texttt{gpt-fr-cased-small} (French), \texttt{german-gpt2} (German), \texttt{gpt2-small-italian} (Italian), \texttt{PULI-GPT-2} (Hungarian), \texttt{norwegian-gpt2} (Norwegian), \texttt{gpt2-chinese-cluecorpussmall} (Mandarin Chinese), and \texttt{japanese-gpt2-medium} (Japanese).}. The use of GPT-2 models is motivated by practical considerations and the need to obtain pre-trained models for each of the eight languages. For each corpus, we derive a fine-tuned version of its corresponding GPT-2 model by further training the model on the corpus dialogues.  To account for the corpus size differences, the number of epochs is adjusted to ensure that the total number of gradient updates is similar across all corpora. 

\begin{figure}[t!]
\vspace{-5mm}\hspace{-2mm}\includegraphics[scale=0.53]{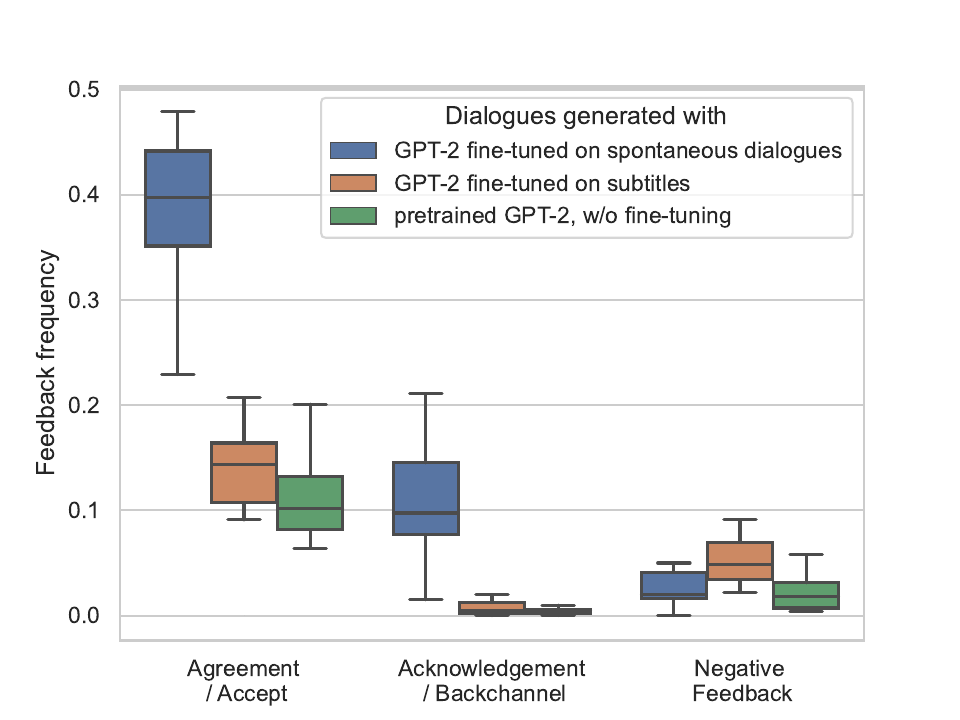}
\caption{Frequency of communicative feedback in synthetic dialogues generated using GPT-2 models, either applied without fine-tuning or after fine-tuning on corpora of spontaneous interactions or subtitles.}
\label{fig:frequency_in_synthetic}
\end{figure}

The GPT-2 models are then employed to produce synthetic dialogues (100 dialogues of about 50 turns per model. For the fine-tuned models, all turns are automatically generated, while for the base models, the following dialogue start is used as context: \emph{Hi! – Hi, how are you? – Fine, and you?} to bias the model towards the generation of dialogues. Finally, the LLM annotator from the previous section is applied on those synthetic dialogues to estimate their frequency of communicative feedback. 

The results are shown in Figure \ref{fig:frequency_in_synthetic}. We observe that the synthetic dialogues generated with the standard GPT-2 models without any further fine-tuning are much closer to the ones derived from subtitles than to those derived from spontaneous interactions when it comes to communicative feedback. This is notably the case for positive and neutral feedback. The occurrence of negative feedback is, however, not as common as in subtitles. Although the above results were obtained here using only GPT-2 pre-trained models, we expect to find similar patterns for other (and more recent) LLMs.

\section{Conclusion and Future Work}
\label{sec:conclusion}

As evidenced in this paper, movie and TV subtitles exhibit notable linguistic differences to actual spontaneous dialogues in the amount and type of conversational feedback they include. Based on a collection of corpora of both spontaneous dialogues and subtitles across eight languages, we provide both lexical statistics and dialogue act estimates derived with a fine-tuned dialogue act tagger. We show that the proportion of conversational feedback is considerably lower in subtitles than in spontaneous dialogues across the corpora included. Furthermore, the type of conversational feedback also differs, as negative feedback is proportionally more frequent in subtitles. This is corroborated by manual annotations of 24 dialogue samples from the selected corpora, and the use of a fine-tuned LLM trained on those annotations. Finally, we also show that dialogues generated from language models are closer to scripted dialogue than real interactions in their use of communicative feedback. Beyond their linguistic interest, these results can provide useful insights for the development of conversational models, as those are often trained on scripted dialogues and might therefore struggle both to understand communicative feedback from the user and to produce such feedback themselves.

\section*{Acknowledgments}
This work was carried out within the \emph{GraphDial} project (\url{https://graphdial.nr.no/}), supported by the Research Council of Norway. The work was initiated during the bilateral PHC-Aurora project {\it French Norwegian Research Effort on Applied Dialogue Modelling}. Laurent Prévot would also like to acknowledge continuous support from the Institute for Language Communication and the Brain (ILCB ANR-16-CONV0002) and the Excellence Initiative of Aix-Marseille University (A*MIDEX AAP-ID-17-46-170301-11.1). Hendrik Buschmeier was supported by the German Research Foundation (DFG) in the Collaborative Research Center TRR 318/1 2021 `Constructing Explainability' (\href{https://gepris.dfg.de/gepris/projekt/438445824}{438445824}). We would like to thank Hiro Yamazaki for helping with the Japanese lexical items and Yu-Lin Chang for annotating the Mandarin data.

\bibliography{bibliography-sigdial2024}

\newpage 
\newpage 

\appendix
\onecolumn

\renewcommand{\thesubsection}{\Alph{subsection}}

\section{Conversational Feedback Lexical Statistics}
\label{sec:appendix}

\subsection*{Cue Word Lists}
In Table \ref{tab:cue_words}, we present the list of cue words used for computing the lexical statistics in Section \ref{sec:lexstats}.\\ \textbf{Content warning}: the lists contain potentially offensive language.
{\small
\begin{longtable}[h]{cc|p{12cm}}\\
\toprule
\textbf{Language} & \textbf{FB} & \textbf{Lexical cues} \\
\midrule
de & + & ja, jaa, jaha, jap, jep, jo, joa, aha, hey, ach, achso, okay, ok, richtig, sicher, verstehe, cool, wow, klar, gut,
definitiv, absolut, genau, natürlich,
ja ja, jaja, ja okay, okay ja, ja genau, ja klar, ja gut, gut okay, ah ja, ja richtig, aber sicher, aber klar, na klar, ich weiß, weiß ich, das stimmt,
du hast recht, sie haben recht, ja genau richtig,
vermutlich, ja vermutlich, aber wirklich \\
& - & nein, nee, nö, niemals, stimmt nicht, das glaube ich nicht,glaube nicht, das glaub ich nicht,glaub nicht, vermutlich nicht\\
& ? & wirklich, bitte, entschuldige, häh, was, wo, warum welchen, welcher, welche, welches, echt, bist du sicher,sind sie sicher\\
& = & mhm, m, mm, hm,ähm,mh,oh,äh \\
\midrule
en & + & yes, yeah, yep, okay, oh, right, alright, good, ok, sure, ah, nice, cool, exactly, absolutely, true, great, oh wow, right right, oh okay, oh yeah, yeah right, um-hum yeah, that's great, yes yes, yeah yeah, uh-huh yeah, that's right, right yeah, oh yes, i see, i know, that right, that's true, that's good, all right, of course, got it, is he, oh that's nice, oh that's good, well that's nice, oh i see, oh that's great, yeah that's true, well that's good, well that's great, right that's right, oh yeah yeah, that sounds good, yeah that's right, yeah yeah yeah, yeah oh yeah, oh yeah oh, well that's true, i guess so, yeah i agree, yeah it is, i think so, oh i know, yeah i know, it really is, it is, i agree, definitely, i do too, you bet, you're right, it does,
i think so too, that's it, i think you're right, i know it, i agree with you, it was, i agree with that, they are, deal, indeed, obviously, clearly, precisely, certainly, no doubt, so do I, I guess so, they really are, it did, they were, they did, me too, to me too, for me too\\
 & - & no, wait, gosh, nope, my goodness, oh no, but um, but uh, stop it, oh my goodness, oh my gosh, wait a minute, oh my god, not really, not much, no way, shit, fuck, oh no\\
 & ? & what, really, oh really, why not, you sure, is that right\\
 & = & um-hum, uh-huh, huh-uh, uh, hum, hm, hey, well, wow, um, huh, mh, mmhmm, m, um-hum um-hum, oh uh-huh, uh-huh uh-huh, um-hum um-hum um-hum, oh, ooh, hmm, mm, mmm\\
\midrule
fr & + & oui, ouais, ok, ah, voilà, bien, daccord, super, parfait, exactement, ah ouais, ouais ouais, et ouais, d'accord, ah oui, oui oui, c'est ça, eh ouais, ah ouais, je sais, très bien, je comprends, bien sûr, ouais ouais ouais, ah ouais ouais, c'est vrai, ah ouais d'accord, ah d'accord, ah ouais OK, ah ouais ok, ah oui oui, ah ben oui, tu m'étonnes, c est bien, sans doute, tout à fait, absolument, vachement, je suis d'accord, moi aussi, c'est vrai, c'est juste, c'est exactement ça\\
 & - & non, putain, pff, si, merde, oh putain, non non, mon dieu, oh mon dieu, je sais pas, non non non, pas trop, pas vraiment, pas possible\\
 & ? & hein, quoi, vraiment, comment ça\\
 & = & ah, mh, euh, oh, han, ben, bon, hm, hum, peut-être, m, mh mh, mh ouais, ah bon, mh mh mh, eh, hé, hey\\
\midrule
no & + & ja, jo, å ja, ok, oi, greit, presis, wow, riktig, sant, nettopp, absolutt, jepp, definitivt, åpenbart, deal, selvfølgelig, sikkert, akkurat, god, bra, helt sikkert, jeg vet, jeg skjønner, helt riktig, det stemmer, klart det, uten tvil, det er riktig, det er greit, det er sant, det er det, jeg er enig, du har rett, det gjør det, jeg tror det, jeg vet det, det var det, det gjør jeg, jeg antar det, det gjorde det, det gjør jeg også, det tror jeg også, jeg tror du har rett, jeg er enig med deg, jeg er enig i det,de er det, de var, det gjorde de, meg også, til meg også, for meg også \\
 & - & nei, faen, javel, herregud, ikke helt, ikke mulig, ikke i det hele tatt\\
 & ? & virkelig, hva, hæ\\
 & = & m, mhm, mh, hmm, mm, mmm, mmhmm, hm, uh-huh, ikke sant\\
 \midrule
 hu & + & igen, tényleg, úgy van, helyes, jogos, igaz, valóban,
    pontosan, tudom, rendben, ok, oké, oksi, okés, okszi, igen az,
    de az, bizony, természetesen, határozottan, feltétlenül, mindenképp, egyetértek, szerintem is, ó igen, hogyne, tényleg az, én is, nekem is, engem is, tőlem is, bennem is, igazad van, naná, mi az hogy, meghiszem azt, biztosra veheted, biztos lehetsz benne, jó, ja, szerintem igen, szerintem is, én is így gondolom, én is úgy gondolom, ennyi, ez az, így van, úgy van, szerintem igazad van, szerintem igazatok van, tudom, jól tudom, egyetértek, az volt, ez volt, de, azok, igen, azok, megegyeztünk, egyértelműen, azt hiszem, kétségtelenül, biztosan, persze, értem, tudod, stimmel, valóban, hát igen, hát dehogynem \\
    & - & nem, nem igazán, nem létezik, a francba, a fenét, ne, a csodát, hogy a csodába, hát nem \\
    & ? & ó tényleg, micsoda, tényleg, miért ne, biztos \\
    & = & aha, hú, ú, ó, óh, hű, ja, mhm, mm, mmm, hmm, hmmm, wow, azta, ejha, nahát, ühüm \\
 \midrule
it & + & ehi, okay, okay, ok, sì, si, vabbè, ecco, perfetto, wow, esatto, certamente, esattamente, assolutamente, sicuramente, decisamente, ovviamente, precisamente, di sicuro, sono d'accordo, concordo, eccellente, grandioso, ottimo, certo, infatti, fantastico, magnifico, naturalmente, giusto, bene, già,lo ben so, ah ah, ah ha, vero, é vero, lo so, lo è, davvero, vero, oh sì, lo è veramente, anch'io, anche io, hai ragione, d'accordo, va bene, benissimo, bello, buono,penso di sì, credo di sì, mi sa di sì, mi pare di sì,anche secondo me, lo penso anch'io, è così, penso che tu abbia ragione, 
penso tu abbia ragione, credo che tu abbia ragione, credo tu abbia ragione, mi sa che hai ragione, sono d'accordo con te, sono d'accordo con voi, lo era, lo è stato, lo è stata, sono d'accordo con ciò, lo sono, senza dubbio, a posto, ci sto, lo sono stati, lo erano, anche a me\\
 & - & oddio, merda, no, non proprio, non molto, non è possibile, cazzo, oh no, macché\\
 & ? & come, davvero, cosa\\
 & = & eh, Mm-hmm, hmm, mmm, mh, eh, mhmh, eh, m, hm, ah, oh, beh, uh-huh, mmh, eeh\\\midrule
 ja & + & \begin{CJK}{UTF8}{min}そう\end{CJK}, \begin{CJK}{UTF8}{min}はい\end{CJK}, \begin{CJK}{UTF8}{min}ええ\end{CJK}, \begin{CJK}{UTF8}{min}そうか\end{CJK},\begin{CJK}{UTF8}{min}はあ\end{CJK}, \begin{CJK}{UTF8}{min}どうぞ\end{CJK}, \begin{CJK}{UTF8}{min}本当\end{CJK}, \begin{CJK}{UTF8}{min}は\end{CJK}, \begin{CJK}{UTF8}{min}あっ\end{CJK}, \begin{CJK}{UTF8}{min}ああ\end{CJK}, \begin{CJK}{UTF8}{min}あ\end{CJK}, \begin{CJK}{UTF8}{min}ね\end{CJK} \\
& = & \begin{CJK}{UTF8}{min}うん\end{CJK}, \begin{CJK}{UTF8}{min}ふーん\end{CJK},\begin{CJK}{UTF8}{min}えっ\end{CJK}, \begin{CJK}{UTF8}{min}へえ\end{CJK}, \begin{CJK}{UTF8}{min}うーん\end{CJK}, \begin{CJK}{UTF8}{min}ふん\end{CJK}, \begin{CJK}{UTF8}{min}え\end{CJK},\begin{CJK}{UTF8}{min}う\end{CJK} \\ 
& - & \begin{CJK}{UTF8}{min}ううん\end{CJK}, \begin{CJK}{UTF8}{min}いいえ\end{CJK},\begin{CJK}{UTF8}{min}いや\end{CJK}, \begin{CJK}{UTF8}{min}いえ\end{CJK}, \begin{CJK}{UTF8}{min}ない\end{CJK}, \begin{CJK}{UTF8}{min}全くない\end{CJK},\begin{CJK}{UTF8}{min}ちょっと\end{CJK} \\
& ? & \begin{CJK}{UTF8}{min}何\end{CJK} \\\midrule
zh & + &  okay, yeah, yes, ok, \begin{CJK*}{UTF8}{gbsn}对\end{CJK*}, \begin{CJK*}{UTF8}{gbsn}哦\end{CJK*}, \begin{CJK*}{UTF8}{gbsn}好\end{CJK*}, \begin{CJK*}{UTF8}{gbsn}是\end{CJK*}, \begin{CJK*}{UTF8}{gbsn}有\end{CJK*}, \begin{CJK*}{UTF8}{gbsn}真的\end{CJK*}, \begin{CJK*}{UTF8}{gbsn}还行\end{CJK*}, \begin{CJK*}{UTF8}{gbsn}當然\end{CJK*}, \begin{CJK*}{UTF8}{gbsn}沒錯\end{CJK*}, \begin{CJK*}{UTF8}{gbsn}太好了\end{CJK*},
          \begin{CJK*}{UTF8}{gbsn}耶\end{CJK*}, \begin{CJK*}{UTF8}{gbsn}行\end{CJK*}, \begin{CJK*}{UTF8}{gbsn}一定\end{CJK*}, \begin{CJK*}{UTF8}{gbsn}没错\end{CJK*}, \begin{CJK*}{UTF8}{gbsn}那好\end{CJK*}, \begin{CJK*}{UTF8}{gbsn}对了\end{CJK*}, \begin{CJK*}{UTF8}{gbsn}真好\end{CJK*}, \begin{CJK*}{UTF8}{gbsn}好啊\end{CJK*}, \begin{CJK*}{UTF8}{gbsn}好吧\end{CJK*}, \begin{CJK*}{UTF8}{gbsn}可以\end{CJK*},
          \begin{CJK*}{UTF8}{gbsn}太棒了\end{CJK*}, \begin{CJK*}{UTF8}{gbsn}太棒了\end{CJK*}, \begin{CJK*}{UTF8}{gbsn}好极了\end{CJK*}, \begin{CJK*}{UTF8}{gbsn}说得对\end{CJK*}, \begin{CJK*}{UTF8}{gbsn}没问题\end{CJK*}, \begin{CJK*}{UTF8}{gbsn}我同意\end{CJK*}, \begin{CJK*}{UTF8}{gbsn}懂了\end{CJK*}, \begin{CJK*}{UTF8}{bsmi}一樣\end{CJK*}, \begin{CJK*}{UTF8}{gbsn}我也是\end{CJK*},
          \begin{CJK*}{UTF8}{gbsn}不错\end{CJK*}, \begin{CJK*}{UTF8}{gbsn}是啊\end{CJK*}, \begin{CJK*}{UTF8}{gbsn}就是这样\end{CJK*}, \begin{CJK*}{UTF8}{gbsn}当然可以\end{CJK*} \\
& - & \begin{CJK*}{UTF8}{gbsn}不\end{CJK*}, \begin{CJK*}{UTF8}{bsmi}沒有\end{CJK*}, \begin{CJK*}{UTF8}{gbsn}不起\end{CJK*}, \begin{CJK*}{UTF8}{gbsn}不是\end{CJK*} \\
& ? & \begin{CJK*}{UTF8}{gbsn}啊\end{CJK*}, \begin{CJK*}{UTF8}{gbsn}是吗\end{CJK*}, \begin{CJK*}{UTF8}{bsmi}什麼\end{CJK*}, \begin{CJK*}{UTF8}{gbsn}什么\end{CJK*}, \begin{CJK*}{UTF8}{gbsn}为什么\end{CJK*} \\
& = & hey, oh, \begin{CJK*}{UTF8}{gbsn}嘿\end{CJK*}, \begin{CJK*}{UTF8}{gbsn}嗯\end{CJK*}, \begin{CJK*}{UTF8}{gbsn}呃\end{CJK*}, \begin{CJK*}{UTF8}{gbsn}哼\end{CJK*}, \begin{CJK*}{UTF8}{gbsn}哈\end{CJK*}, \begin{CJK*}{UTF8}{gbsn}嘘\end{CJK*}, \begin{CJK*}{UTF8}{gbsn}喔\end{CJK*}, \begin{CJK*}{UTF8}{gbsn}呵呵\end{CJK*}, \begin{CJK*}{UTF8}{gbsn}噢\end{CJK*}, \begin{CJK*}{UTF8}{gbsn}哇\end{CJK*}, \begin{CJK*}{UTF8}{gbsn}哦\end{CJK*}, \begin{CJK*}{UTF8}{gbsn}哟\end{CJK*}, \begin{CJK*}{UTF8}{gbsn}咦\end{CJK*} \\
[2mm]\bottomrule \\
\caption{Lists of cue phrases employed in the lexical overview of Section \ref{sec:lexstats}. We distinguish between four core categories of feedback, namely positive feedback/acknowledgment (\textbf{+}), neutral/continuer (\textbf{=}), negative (\textbf{-}), and clarification request (\textbf{?}).} \label{tab:cue_words}
\end{longtable}}

\subsection*{Lexical Statistics plots}

\begin{figure*}[hbt]
     \centering
     \begin{subfigure}[b]{0.325\linewidth}
         \centering
         \includegraphics[trim=6mm 2mm 0 7mm,clip,scale=0.28]{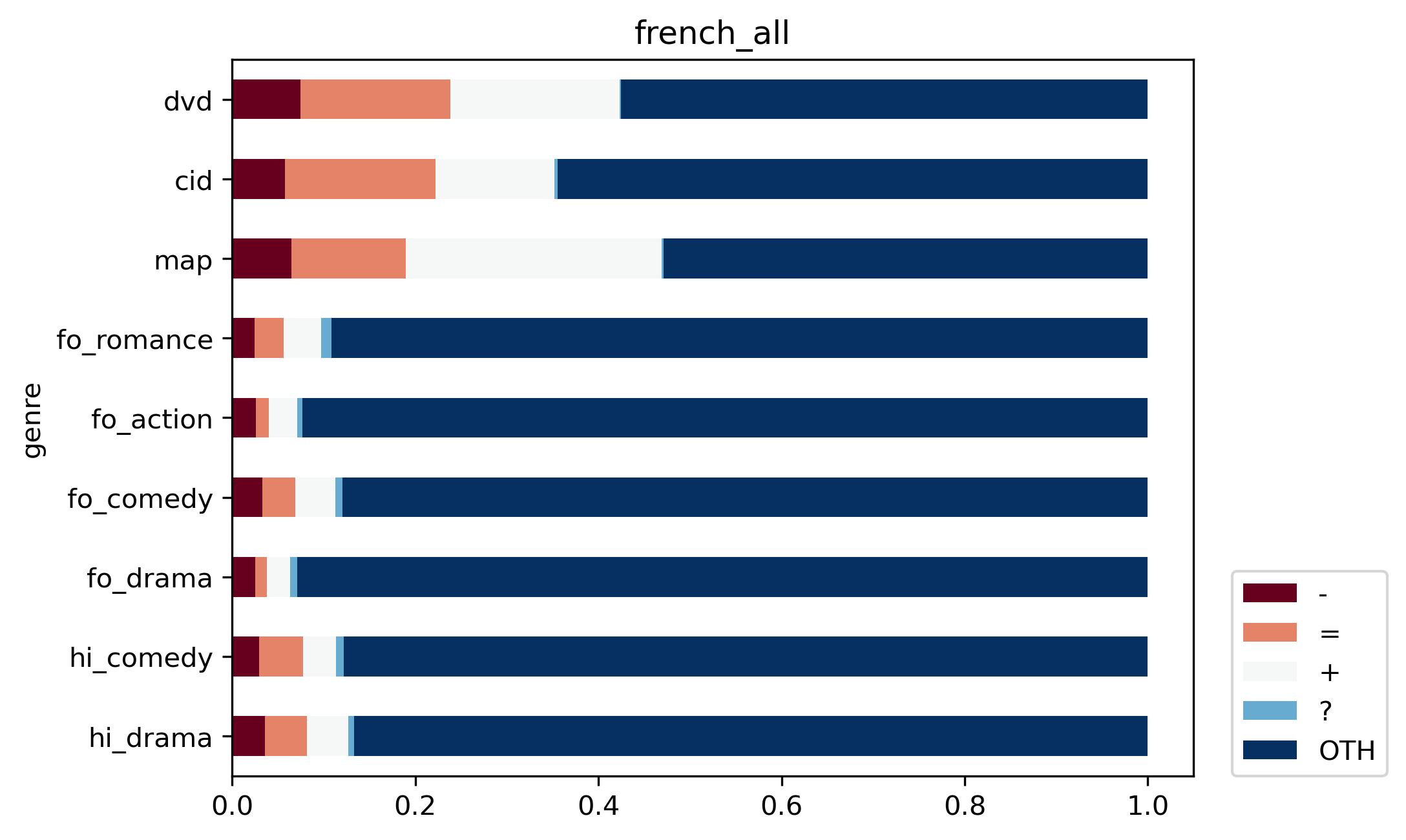}
         \caption{Utterances type}
         \label{fig:fr_utts}
     \end{subfigure}
     \!
     \begin{subfigure}[b]{0.325\linewidth}
         \centering
         \includegraphics[trim=6mm 2mm 0 7mm,clip,scale=0.28]{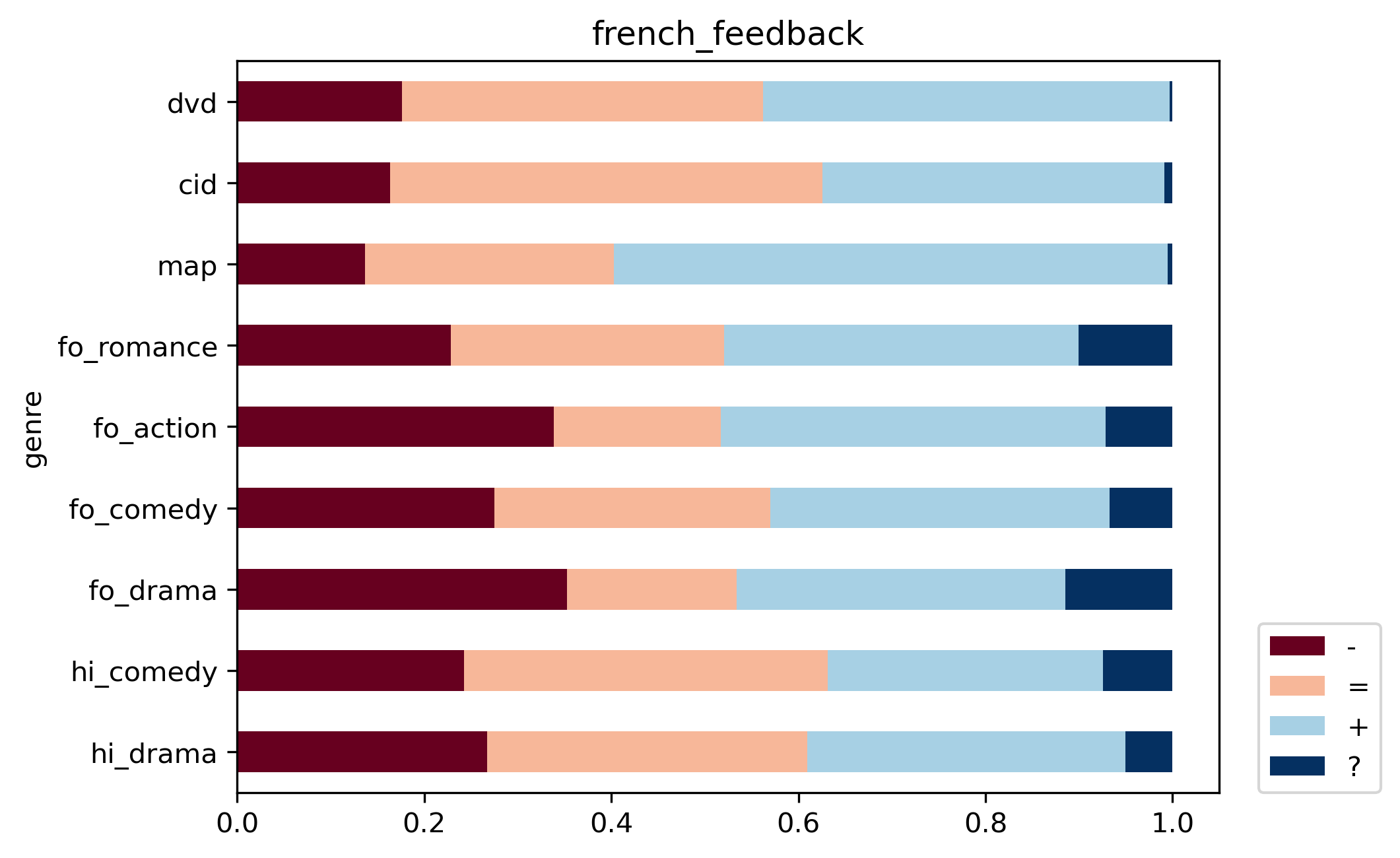}
         \caption{Feedback types}
         \label{fig:fr_feed}
     \end{subfigure}
     \!
     \begin{subfigure}[b]{0.325\linewidth}
         \centering
         \includegraphics[trim=6mm 2mm 0 7mm,clip,scale=0.28]{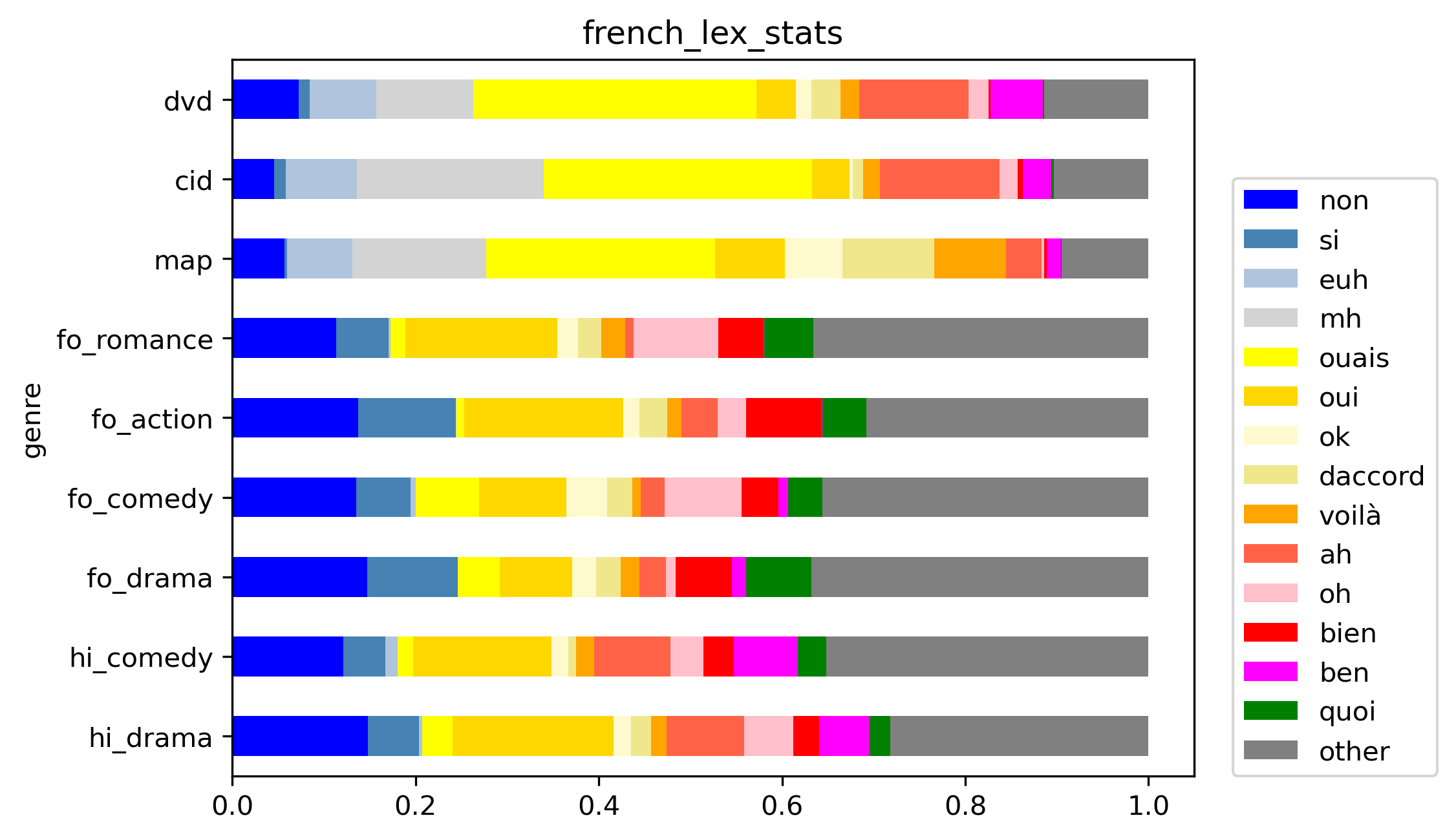}
         \caption{Lexical items}
         \label{fig:fr_lex}
     \end{subfigure}
        \caption{French across genres (rule-based, based on cue word lists).}
        \label{fig:lextstats_fr}
\end{figure*}

\begin{figure*}
     \centering
     \begin{subfigure}[b]{0.325\linewidth}
         \centering
         \includegraphics[trim=6mm 2mm 0 7mm,clip,scale=0.28]{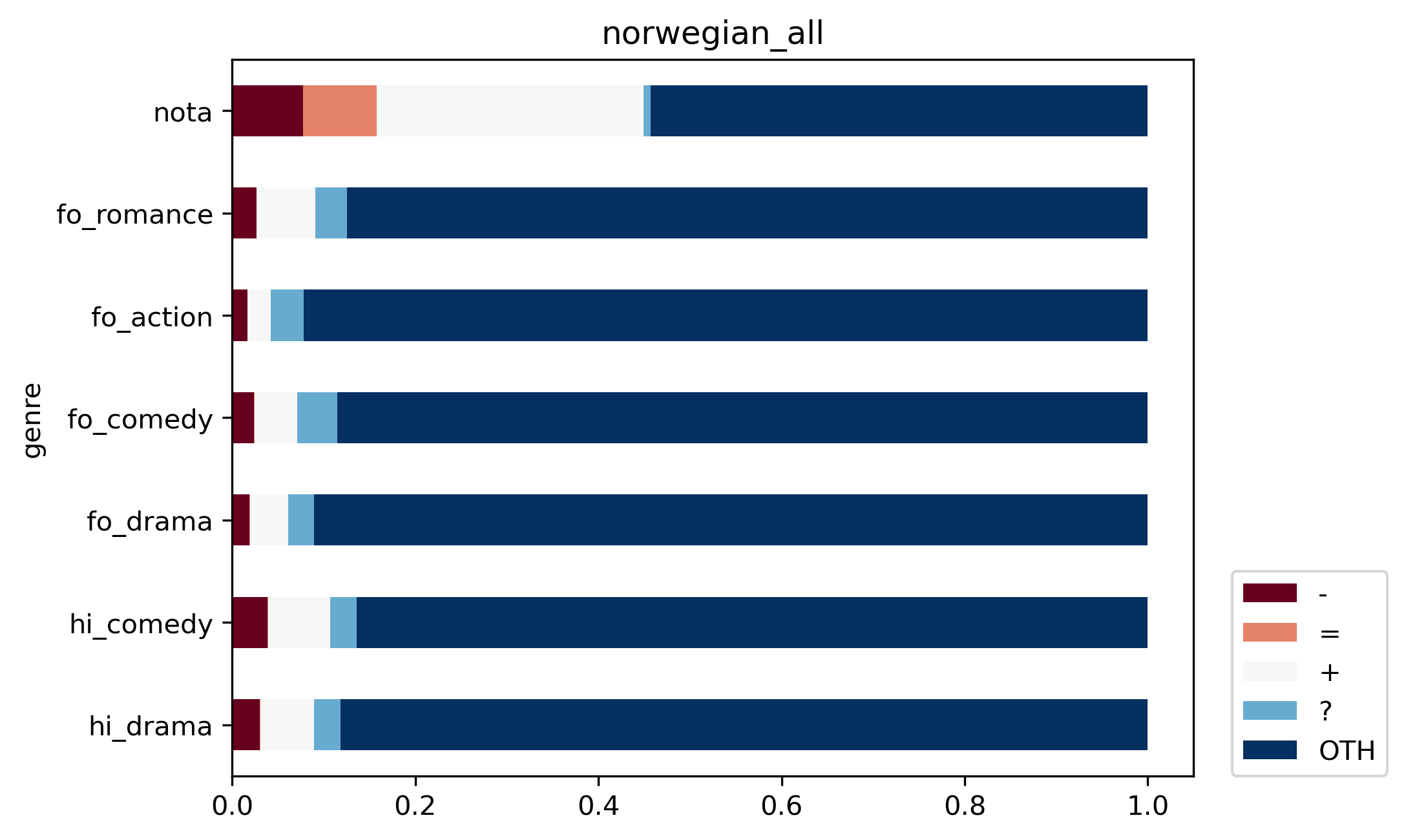}
         \caption{Utterances type}
         \label{fig:no_utts}
     \end{subfigure}
     \hfill
     \begin{subfigure}[b]{0.325\linewidth}
         \centering
         \includegraphics[trim=6mm 2mm 0 7mm,clip,scale=0.28]{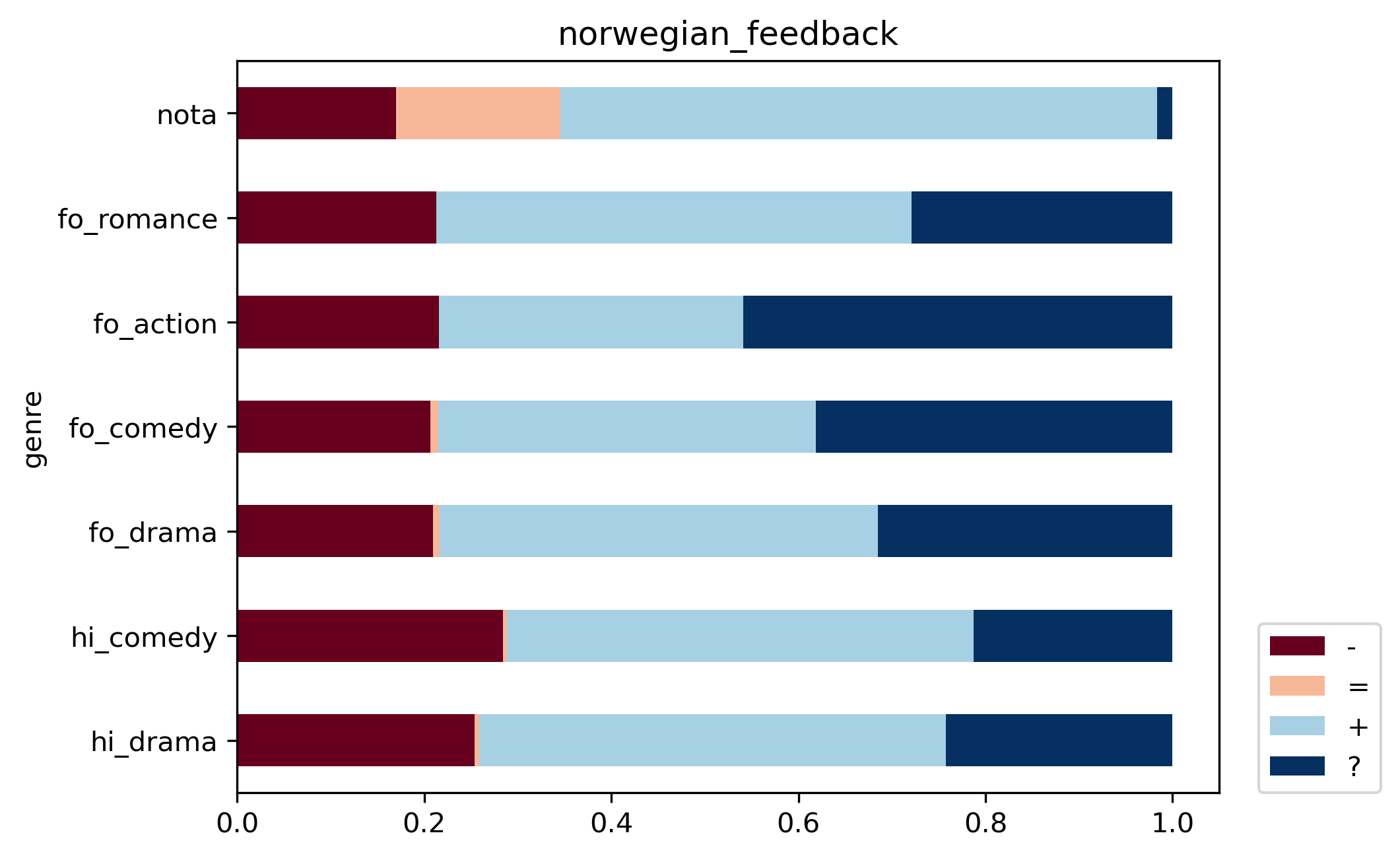}
         \caption{Feedback types}
         \label{fig:no_feed}
     \end{subfigure}
     \hfill
     \begin{subfigure}[b]{0.325\linewidth}
         \centering
         \includegraphics[trim=6mm 2mm 0 7mm,clip,scale=0.28]{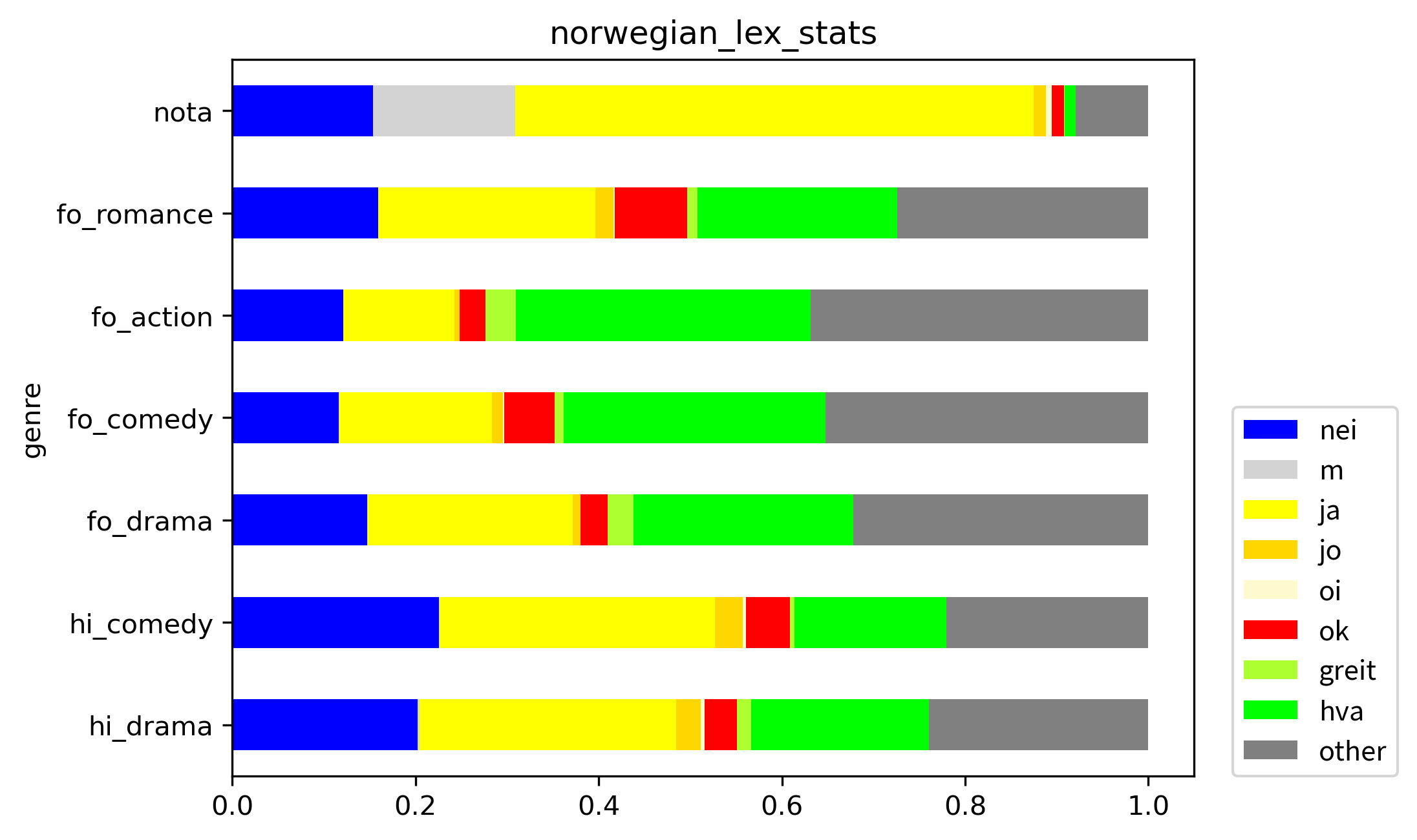}
         \caption{Lexical items}
         \label{fig:no_lex}
     \end{subfigure}
        \caption{Norwegian across genres (rule-based, based on cue word lists).}
        \label{fig:lextstats_no}
\end{figure*}

\begin{figure*}
     \centering
     \begin{subfigure}[b]{0.325\linewidth}
         \centering
         \includegraphics[trim=6mm 2mm 0 7mm,clip,scale=0.28]{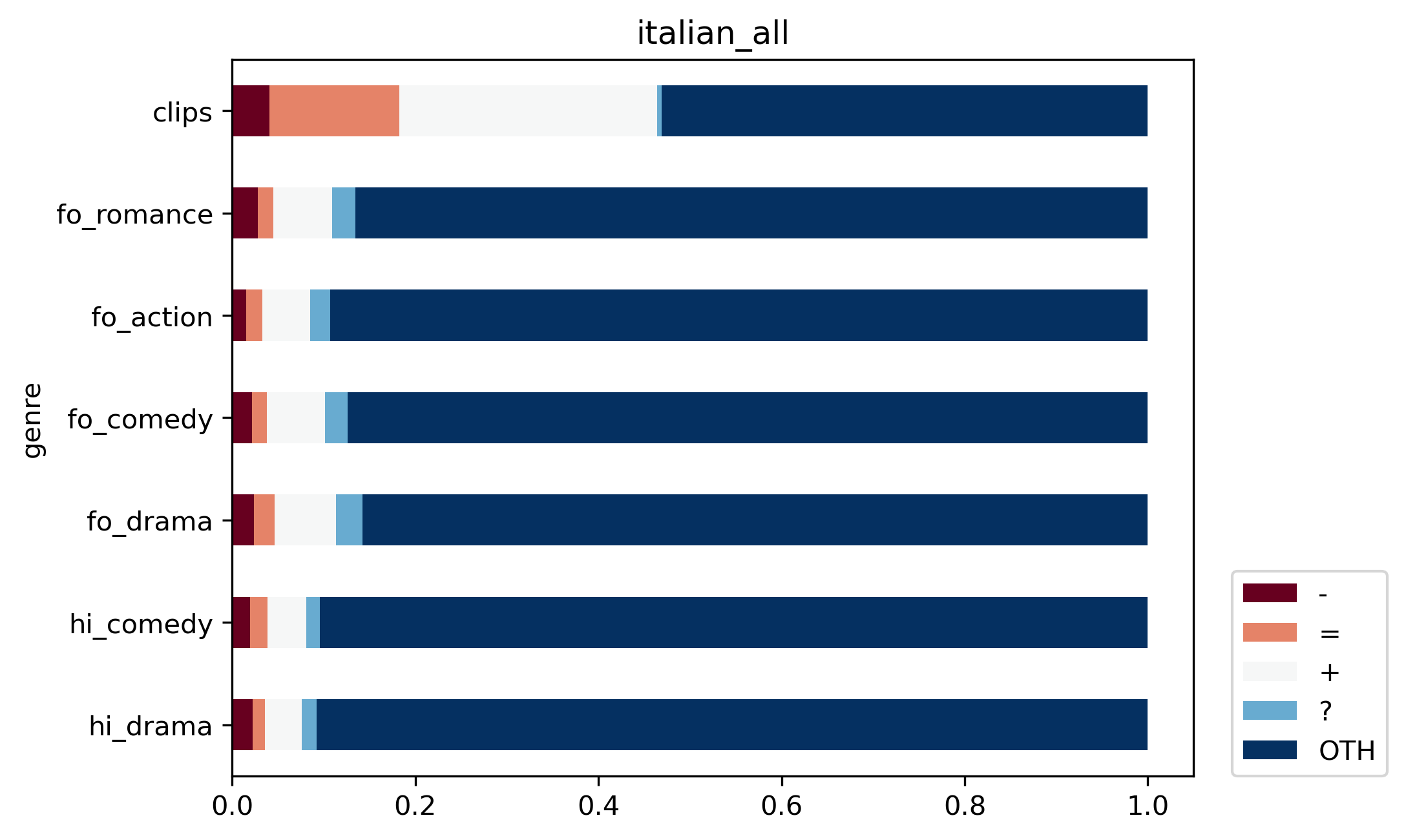}
         \caption{Utterances type}
         \label{fig:it_utts}
     \end{subfigure}
     \hfill
     \begin{subfigure}[b]{0.325\linewidth}
         \centering
         \includegraphics[trim=6mm 2mm 0 7mm,clip,scale=0.28]{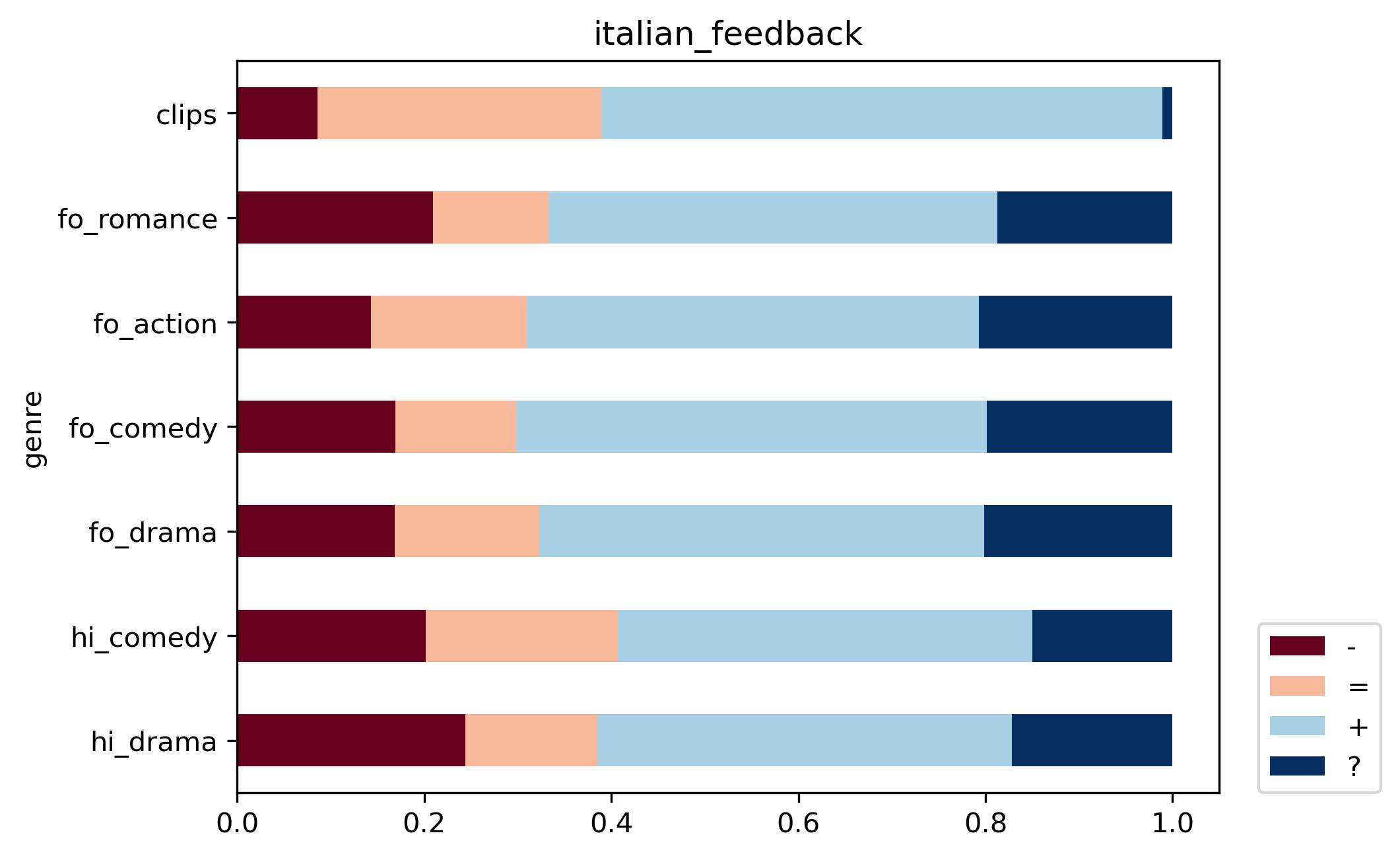}
         \caption{Feedback types}
         \label{fig:it_feed}
     \end{subfigure}
     \hfill
     \begin{subfigure}[b]{0.325\linewidth}
         \centering
         \includegraphics[trim=6mm 2mm 0 7mm,clip,scale=0.28]{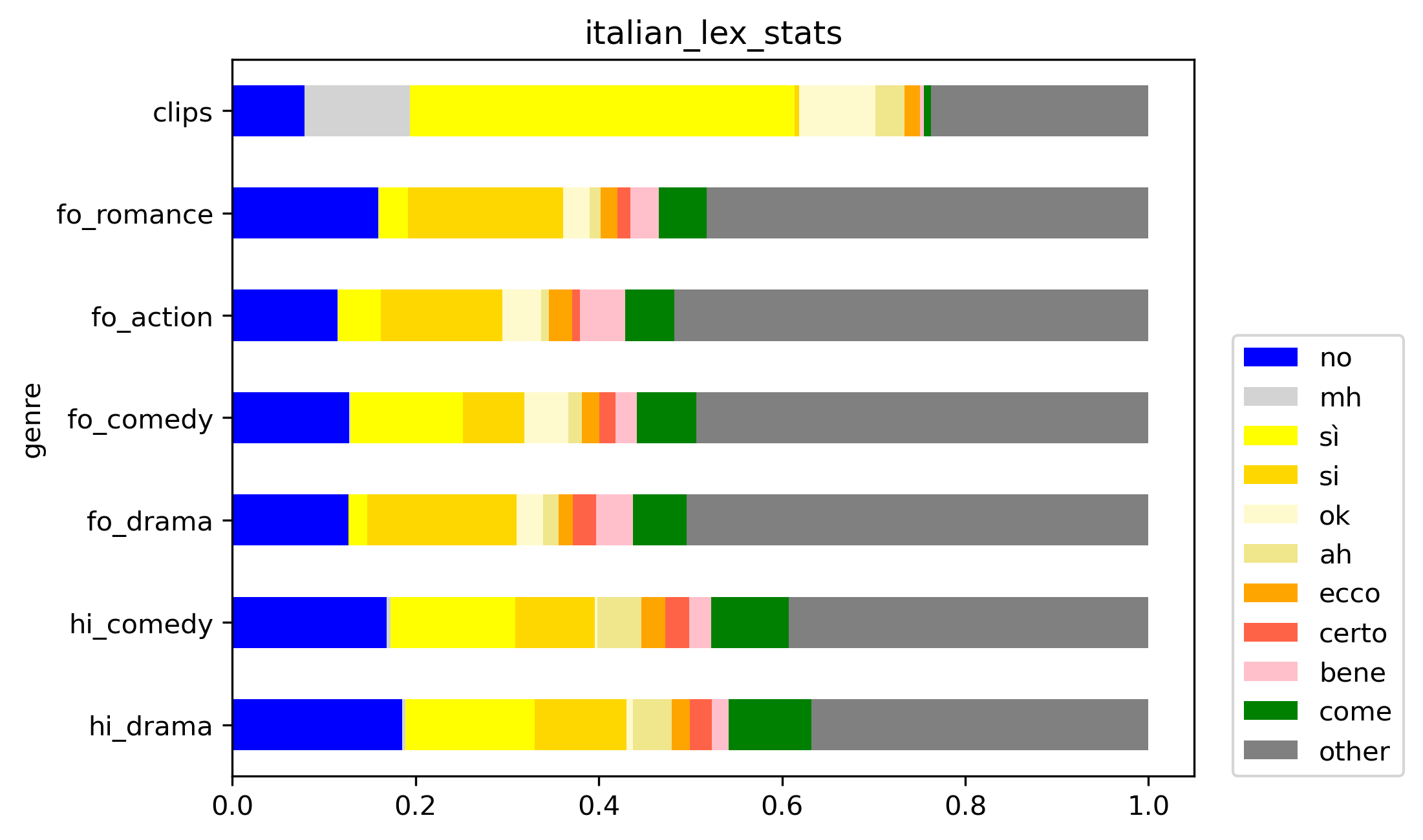}
         \caption{Lexical items}
         \label{fig:it_lex}
     \end{subfigure}
        \caption{Italian across genres (rule-based, based on cue word lists).}
        \label{fig:lextstats_it}
\end{figure*}

\begin{figure*}
     \centering
     \begin{subfigure}[b]{0.325\linewidth}
         \centering
         \includegraphics[trim=6mm 2mm 0 7mm,clip,scale=0.28]{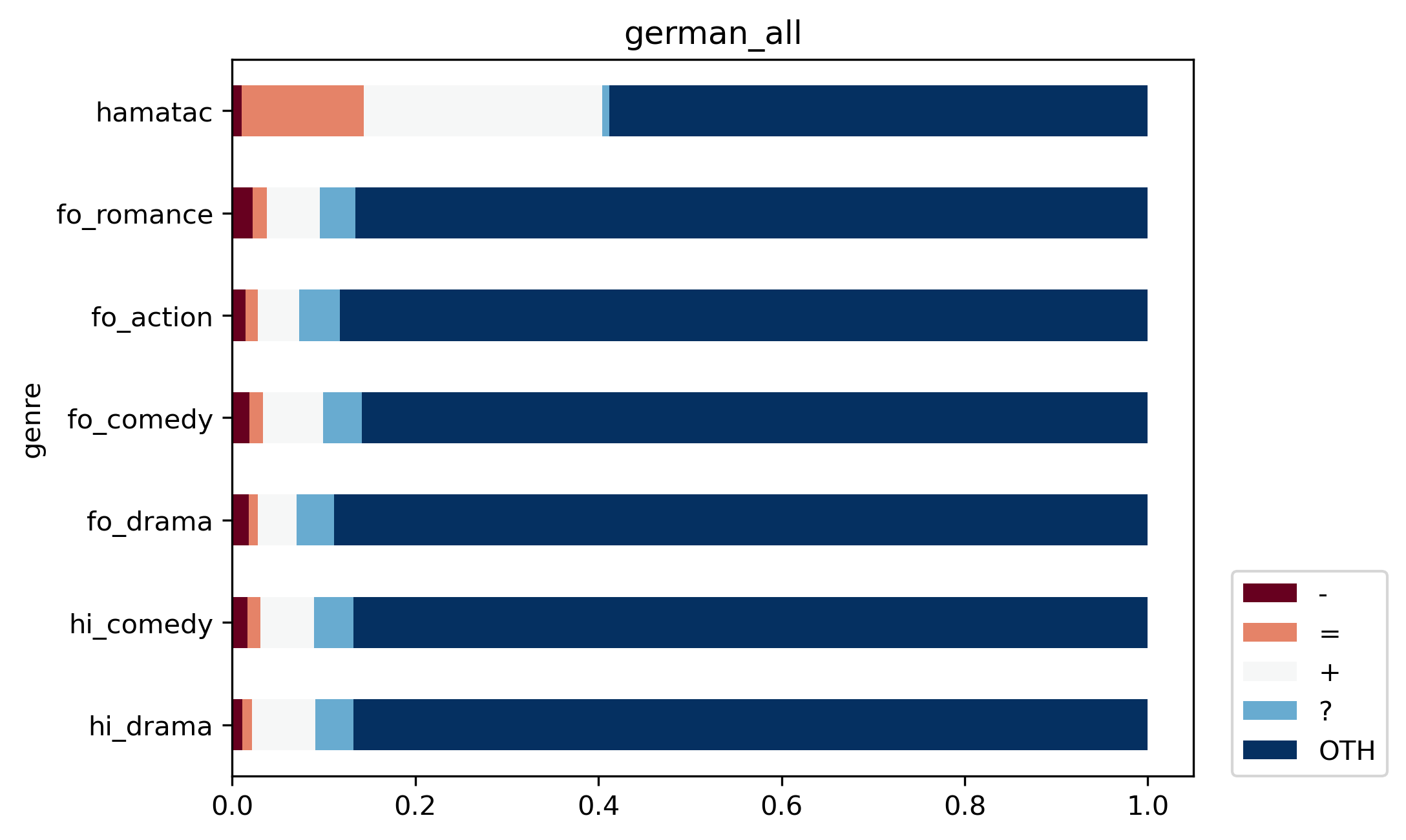}
         \caption{Utterances type}
         \label{fig:de_utts}
     \end{subfigure}
     \hfill
     \begin{subfigure}[b]{0.325\linewidth}
         \centering
         \includegraphics[trim=6mm 2mm 0 7mm,clip,scale=0.28]{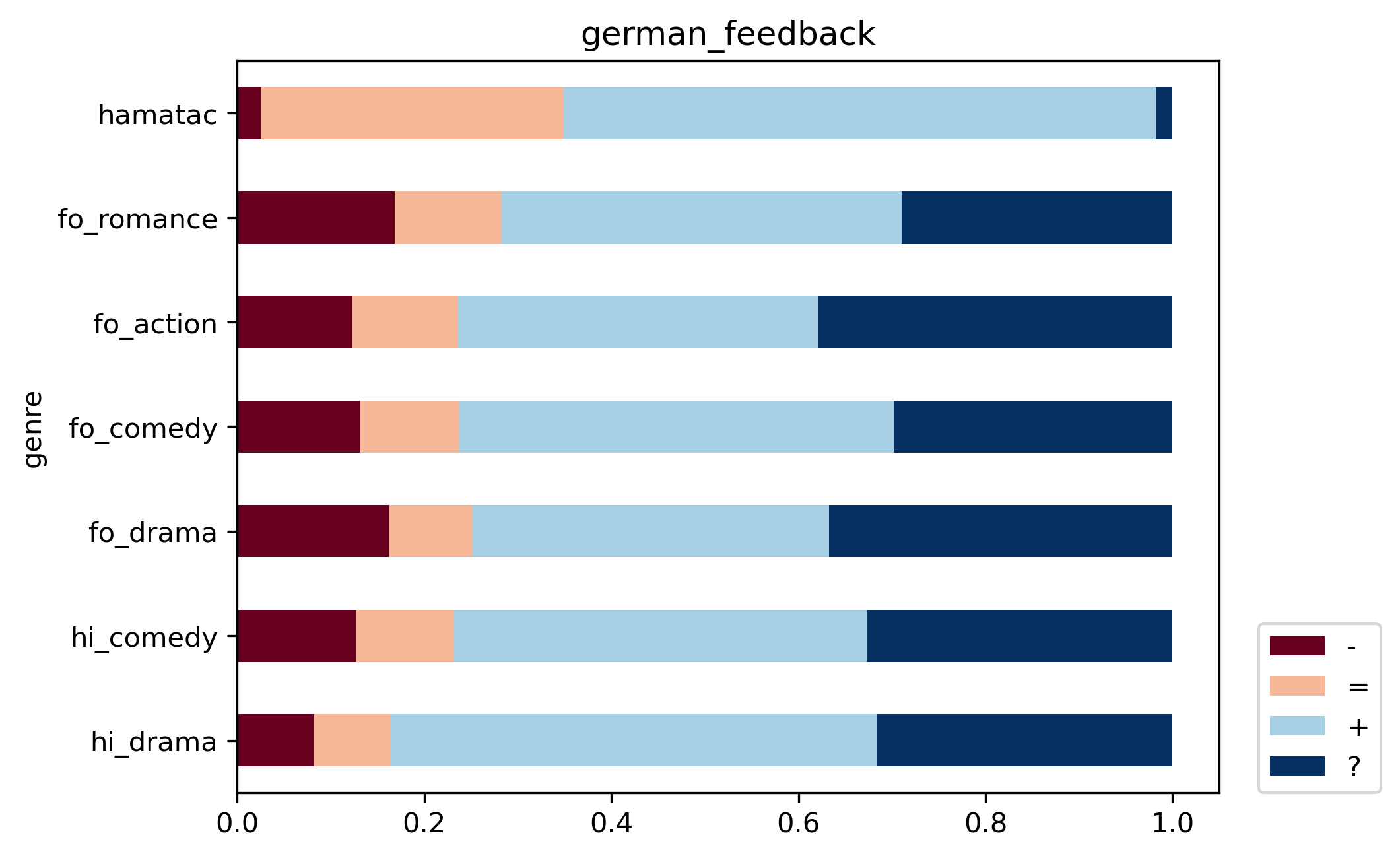}
         \caption{Feedback types}
         \label{fig:de_feed}
     \end{subfigure}
     \hfill
     \begin{subfigure}[b]{0.325\linewidth}
         \centering
         \includegraphics[trim=6mm 2mm 0 7mm,clip,scale=0.28]{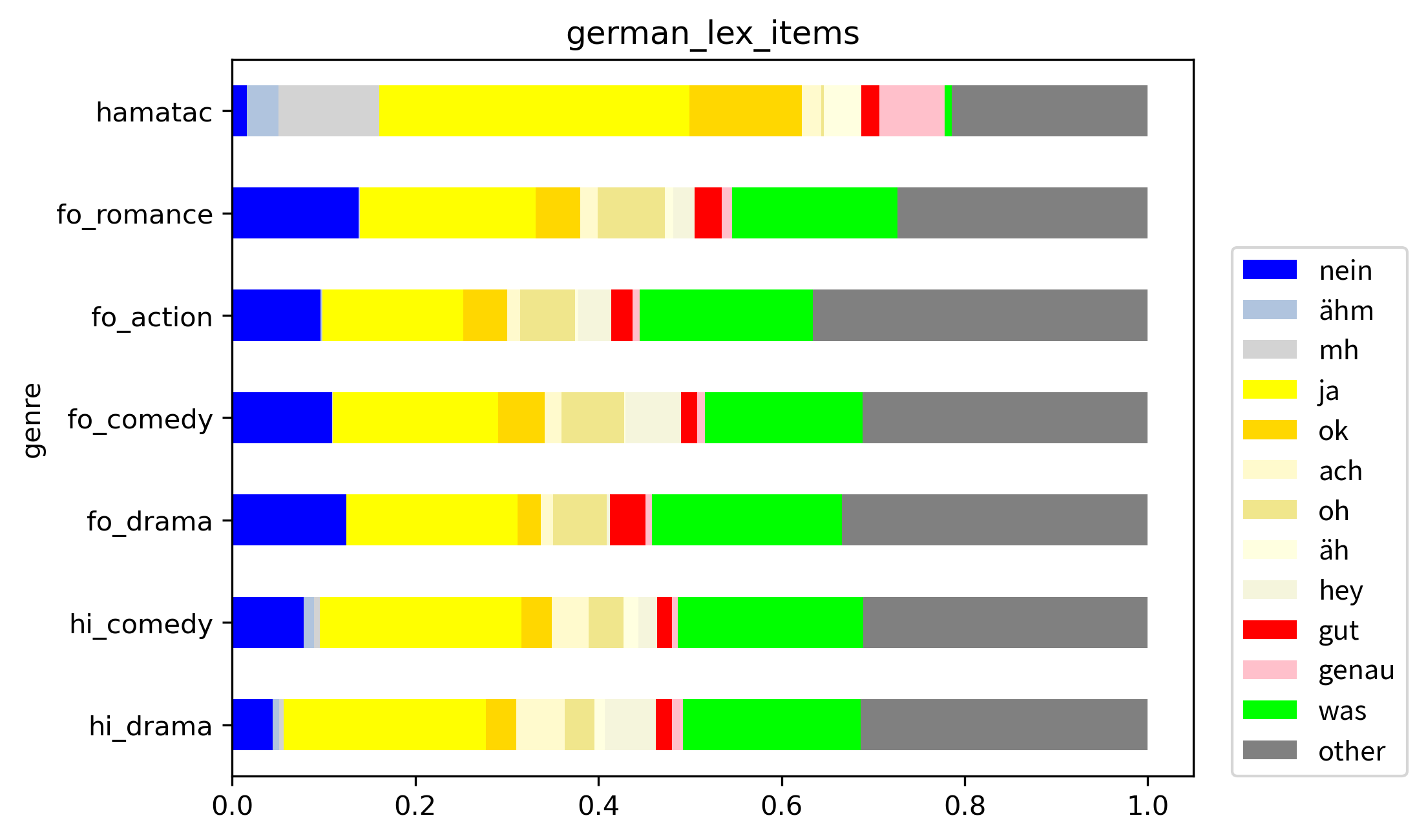}
         \caption{Lexical items}
         \label{fig:de_lex}
     \end{subfigure}
        \caption{German across genres (rule-based, based on cue word lists).}
        \label{fig:lextstats_de}
\end{figure*}

\begin{figure*}
     \centering
     \begin{subfigure}[b]{0.325\linewidth}
         \centering
         \includegraphics[trim=6mm 2mm 0 7mm,clip,scale=0.28]{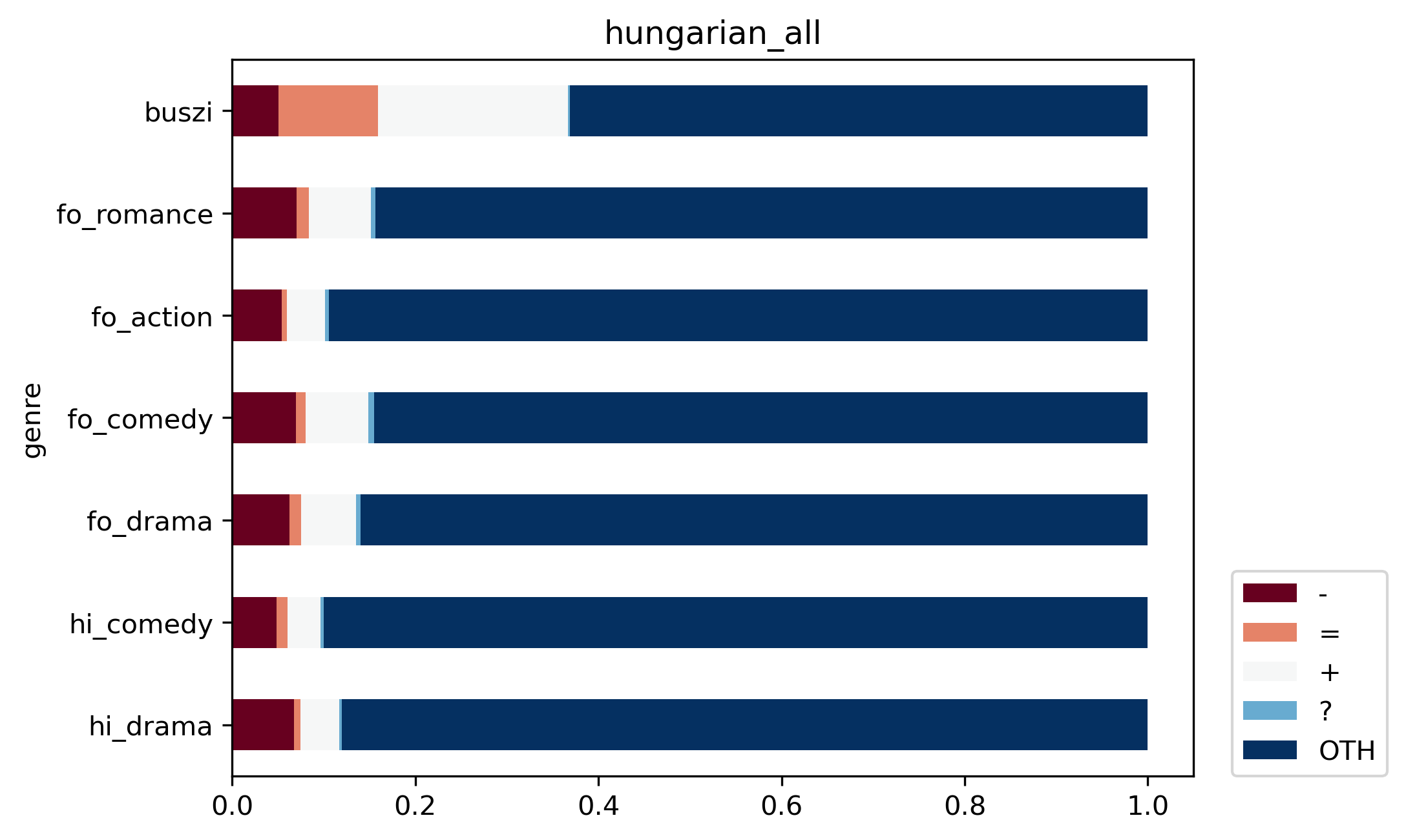}
         \caption{Utterances type}
         \label{fig:hu_utts}
     \end{subfigure}
     \hfill
     \begin{subfigure}[b]{0.325\linewidth}
         \centering
         \includegraphics[trim=6mm 2mm 0 7mm,clip,scale=0.28]{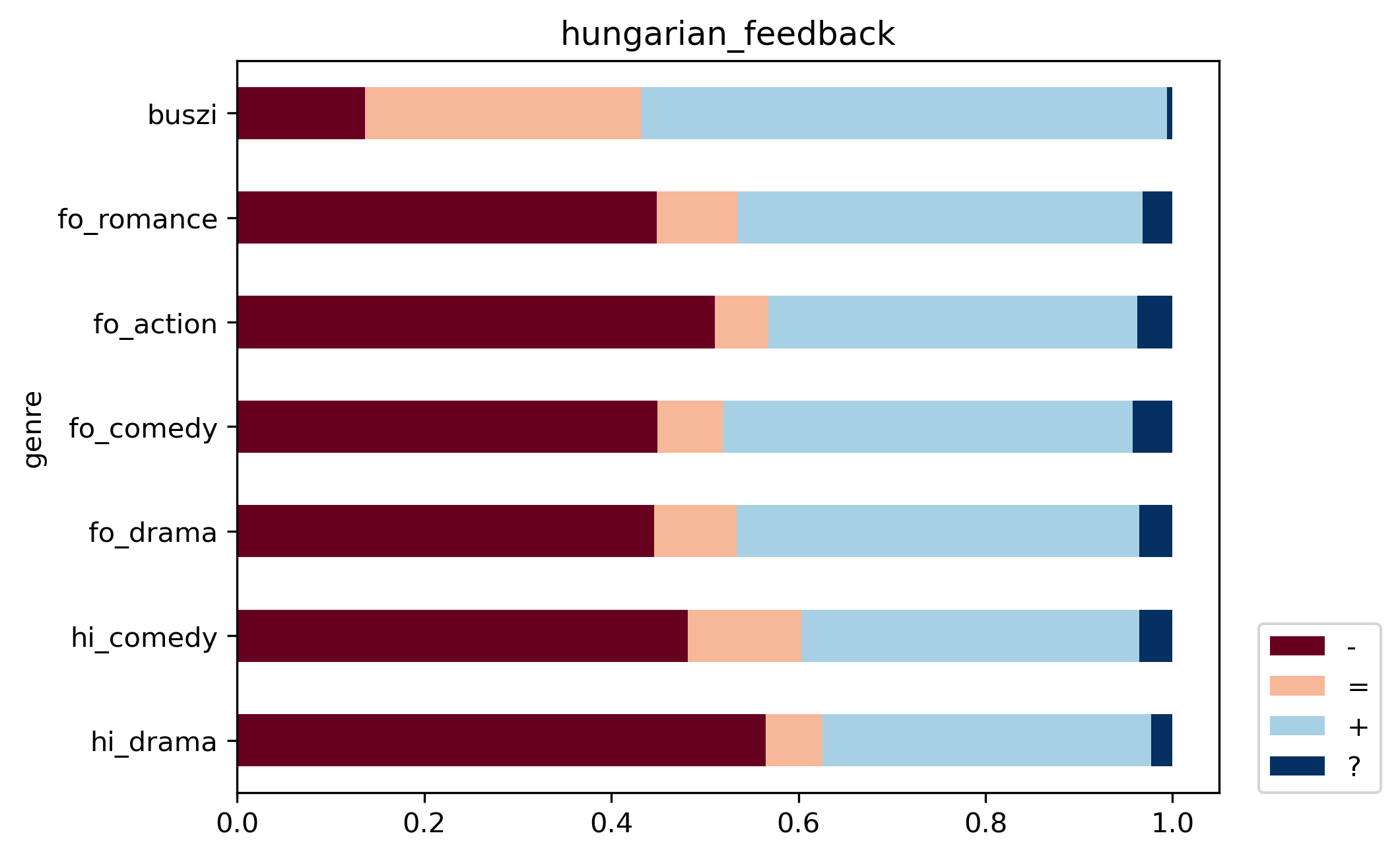}
         \caption{Feedback types}
         \label{fig:hu_feed}
     \end{subfigure}
     \hfill
     \begin{subfigure}[b]{0.325\linewidth}
         \centering
         \includegraphics[trim=6mm 2mm 0 7mm,clip,scale=0.28]{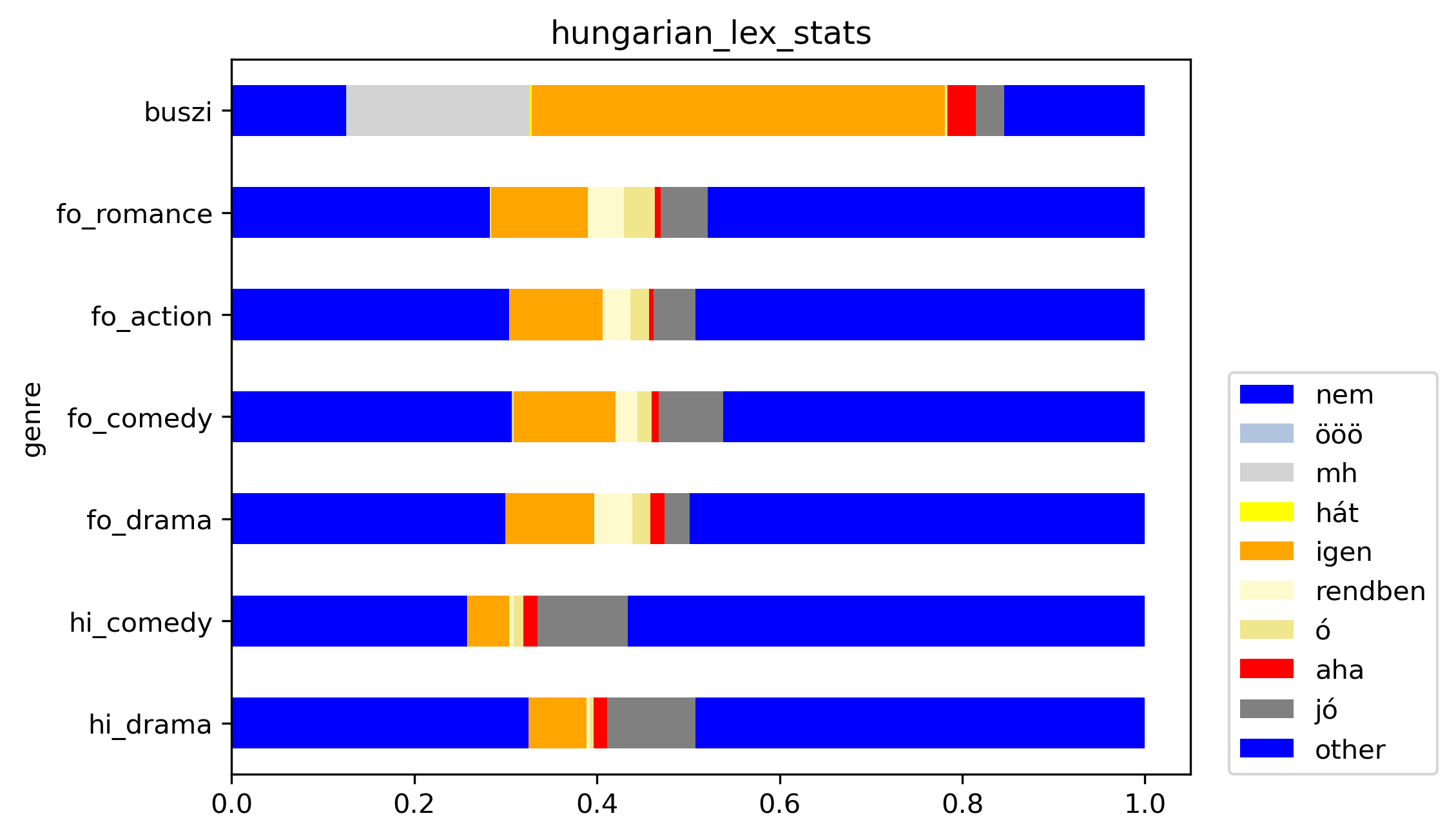}
         \caption{Lexical items}
         \label{fig:hu_lex}
     \end{subfigure}
        \caption{Hungarian across genres (rule-based, based on cue word lists).}
        \label{fig:lextstats_hu}
\end{figure*}

\begin{figure*}
     \centering
     \begin{subfigure}[b]{0.325\linewidth}
         \centering
         \includegraphics[trim=6mm 2mm 0 7mm,clip,scale=0.28]{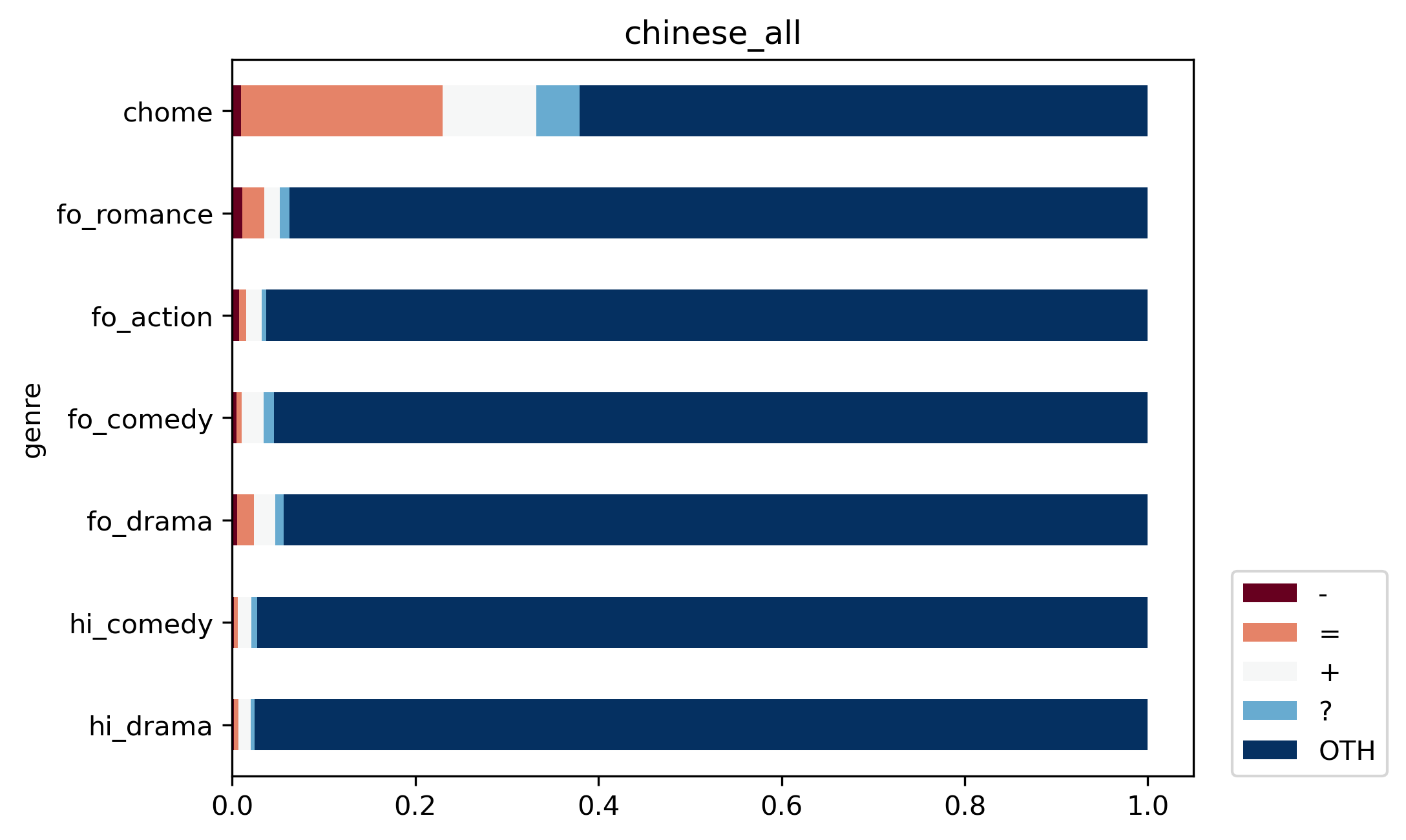}
         \caption{Utterances type}
         \label{fig:cn_utts_v2}
     \end{subfigure}
     \hfill
     \begin{subfigure}[b]{0.325\linewidth}
         \centering
         \includegraphics[trim=6mm 2mm 0 7mm,clip,scale=0.28]{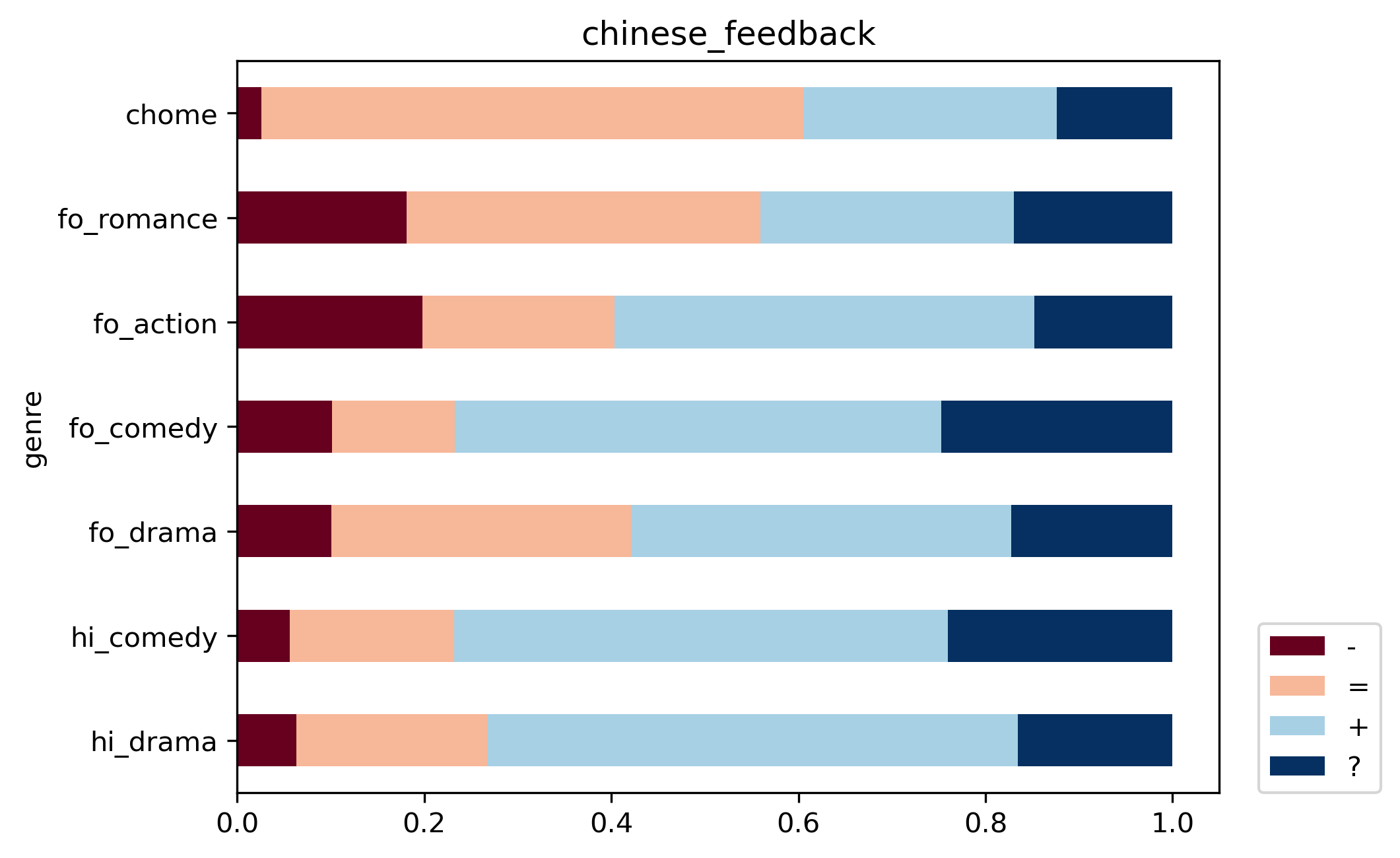}
         \caption{Feedback types}
         \label{fig:cn_feed_v2}
     \end{subfigure}
     \hfill
     \begin{subfigure}[b]{0.325\linewidth}
         \centering
         \includegraphics[trim=6mm 2mm 0 7mm,clip,scale=0.28]{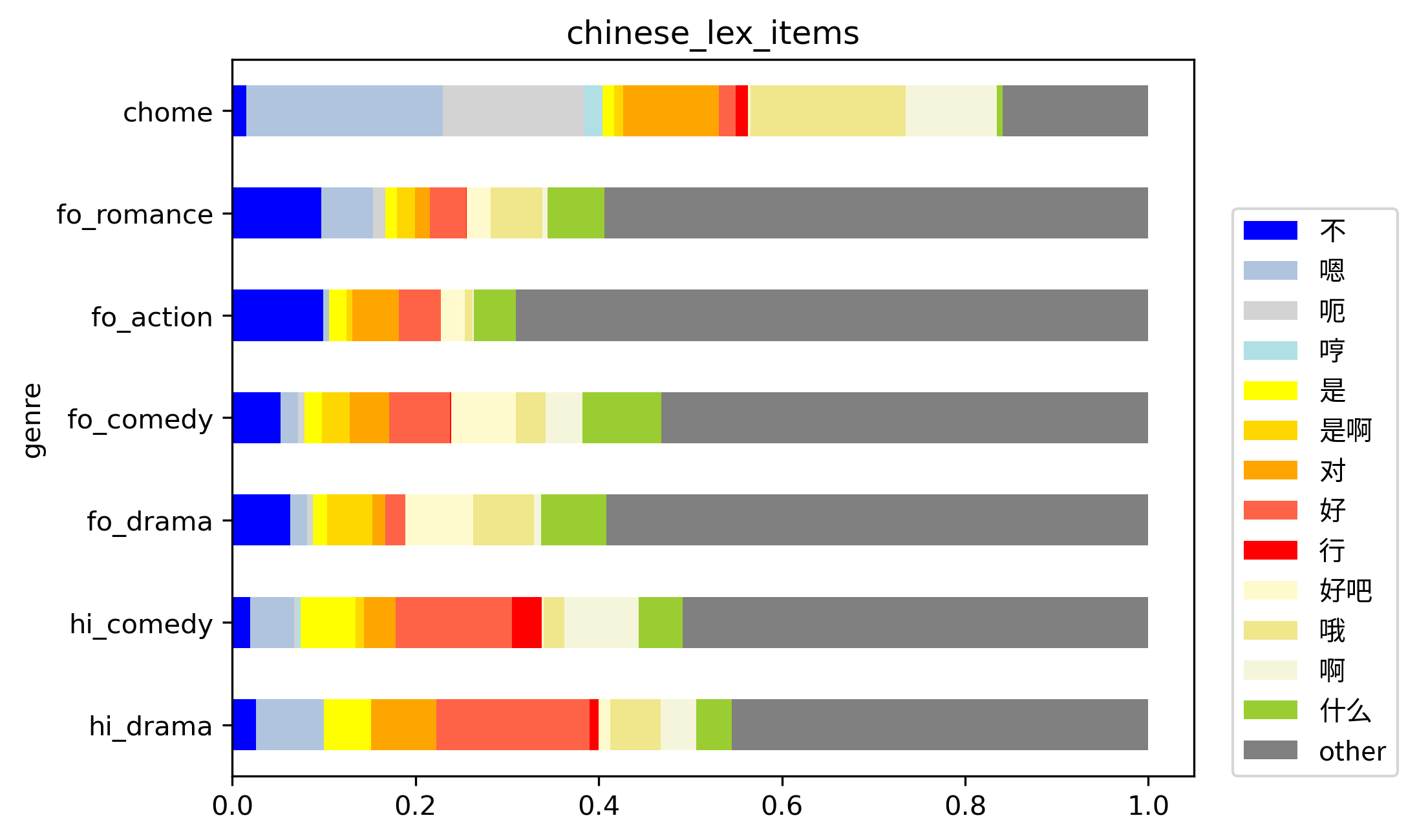}
         \caption{Lexical items}
         \label{fig:cn_lex}
     \end{subfigure}
        \caption{Mandarin Chinese across genres (rule-based, based on cue word lists).}
        \label{fig:lextstats_cn_v2}
\end{figure*}

\begin{figure*}
     \centering
     \begin{subfigure}[b]{0.325\linewidth}
         \centering
         \includegraphics[trim=6mm 2mm 0 7mm,clip,scale=0.28]{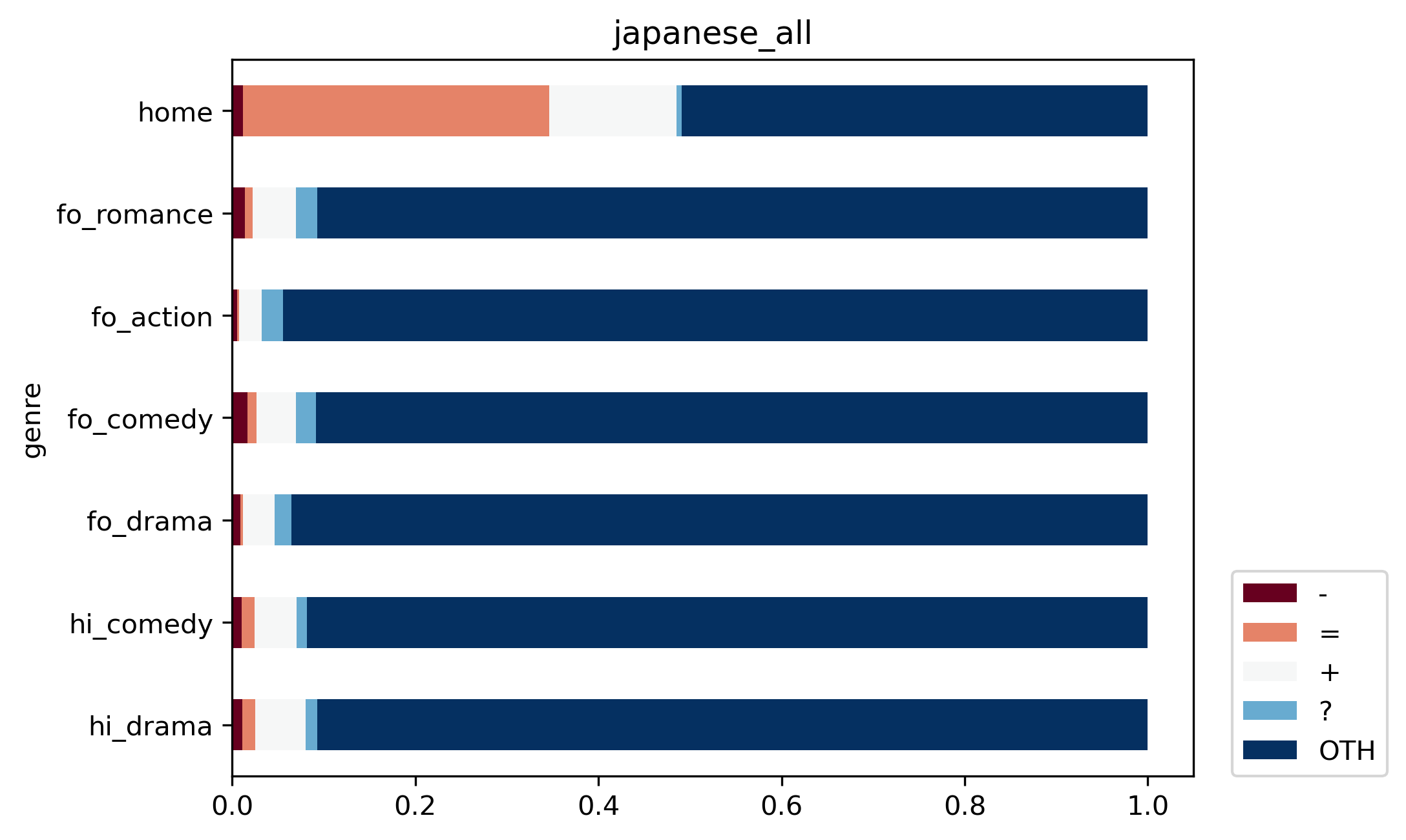}
         \caption{Utterances type}
         \label{fig:ja_utts}
     \end{subfigure}
     \hfill
     \begin{subfigure}[b]{0.325\linewidth}
         \centering
         \includegraphics[trim=6mm 2mm 0 7mm,clip,scale=0.28]{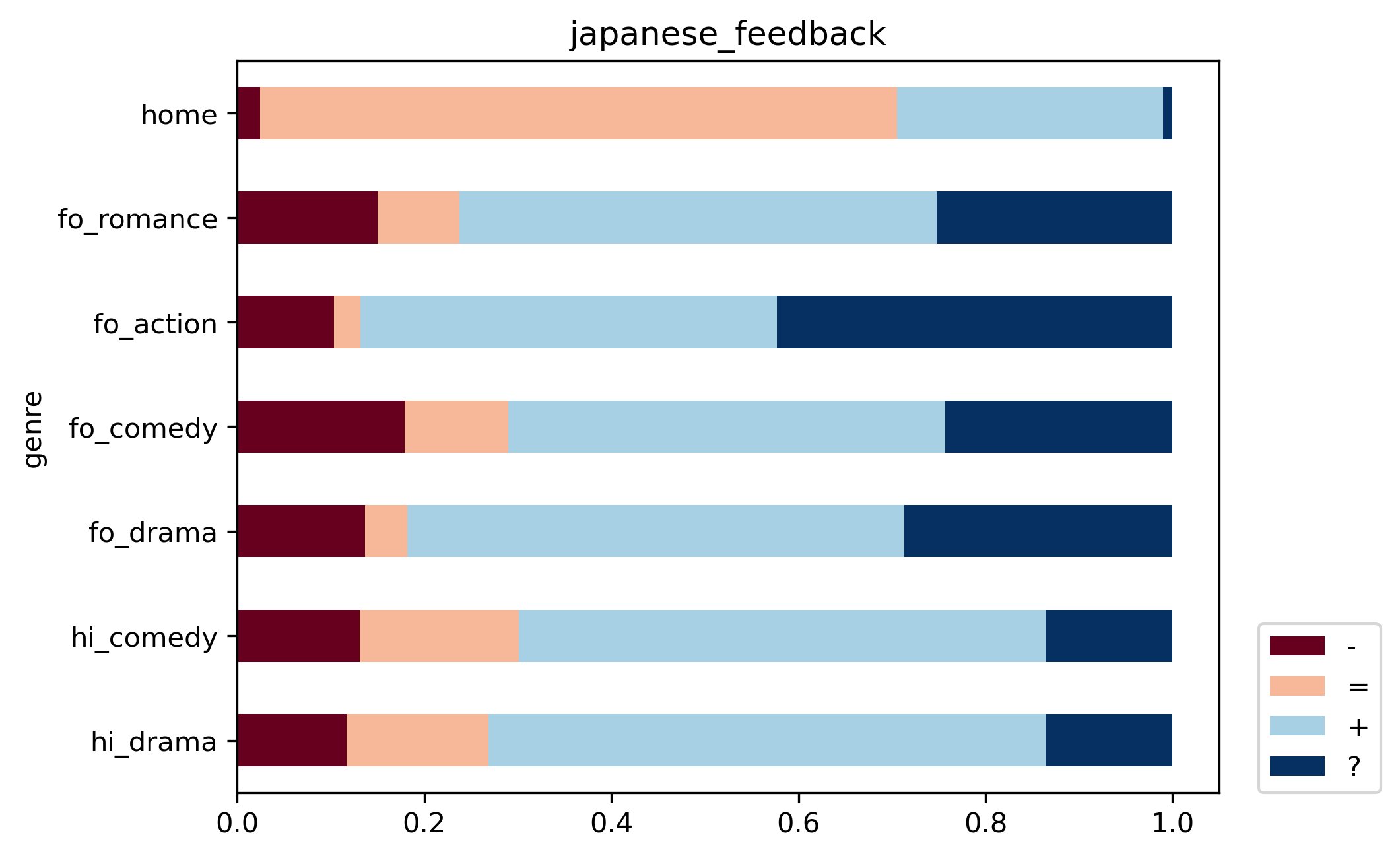}
         \caption{Feedback types}
         \label{fig:ja_feed}
     \end{subfigure}
     \hfill
     \begin{subfigure}[b]{0.325\linewidth}
         \centering
         \includegraphics[trim=6mm 2mm 0 7mm,clip,scale=0.28]{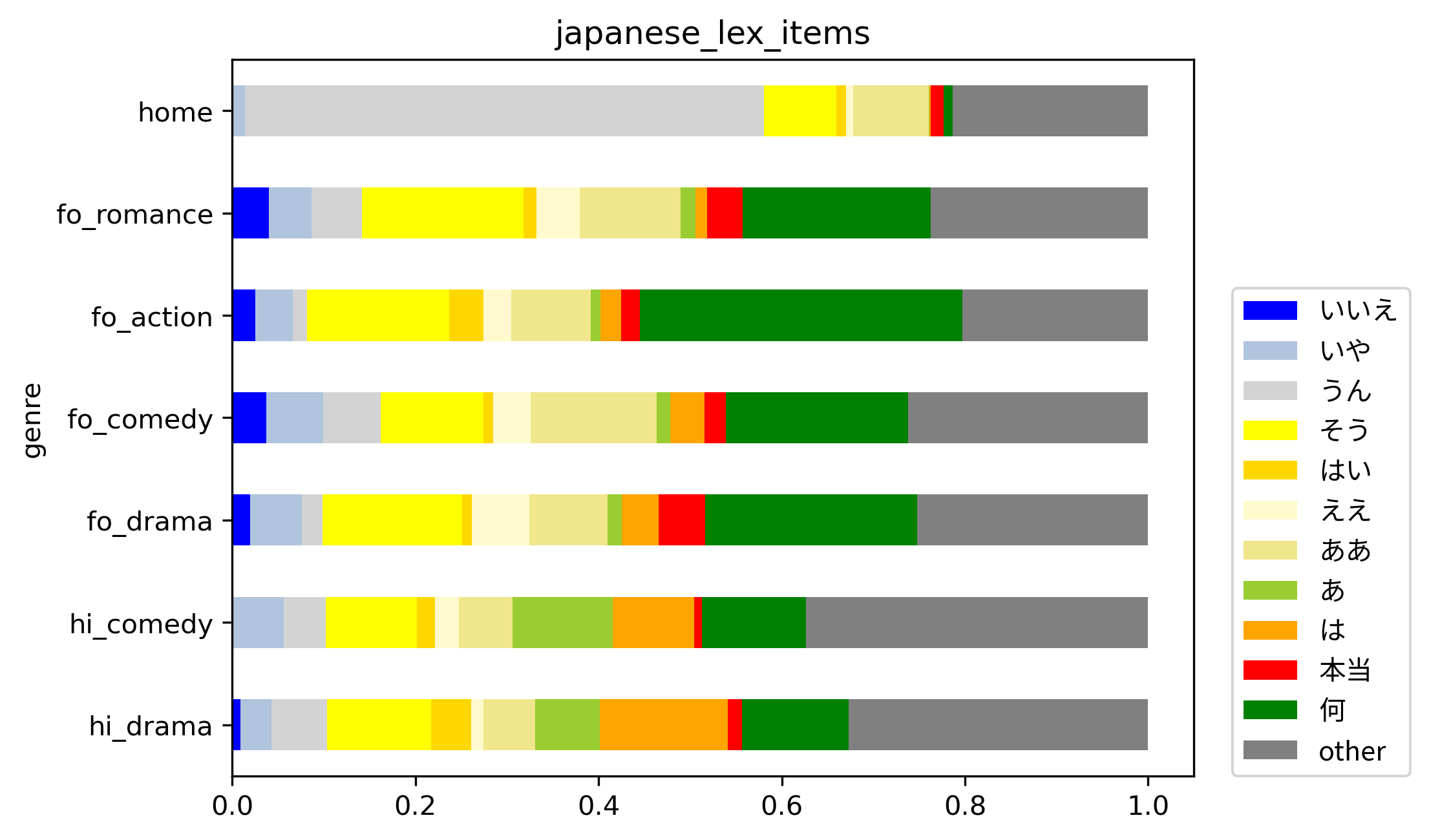}
         \caption{Lexical items}
         \label{fig:ja_lex}
     \end{subfigure}
        \caption{Japanese across genres (rule-based, based on cue word lists).}
        \label{fig:lextstats_ja}
\end{figure*}

\begin{figure*}
     \centering
     \begin{subfigure}[b]{0.49\linewidth}
         \centering
         \includegraphics[trim=6mm 2mm 0 7mm,clip,scale=0.43]{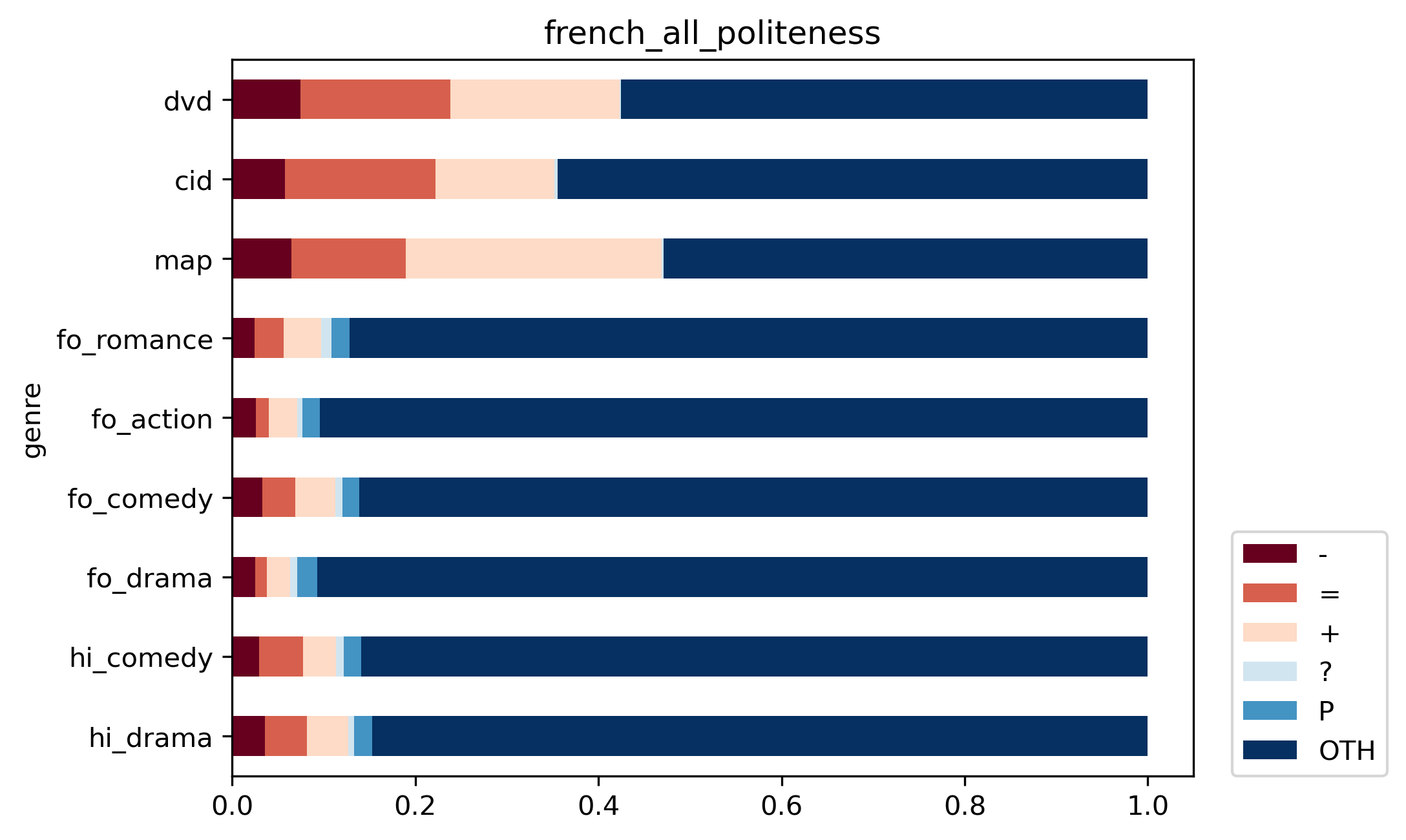}
         \caption{French including politeness ($P$) keywords}
         \label{fig:fr_politeness}
     \end{subfigure}
     \hfill
     \begin{subfigure}[b]{0.49\linewidth}
         \centering
         \includegraphics[trim=6mm 2mm 0 7mm,clip,scale=0.43]{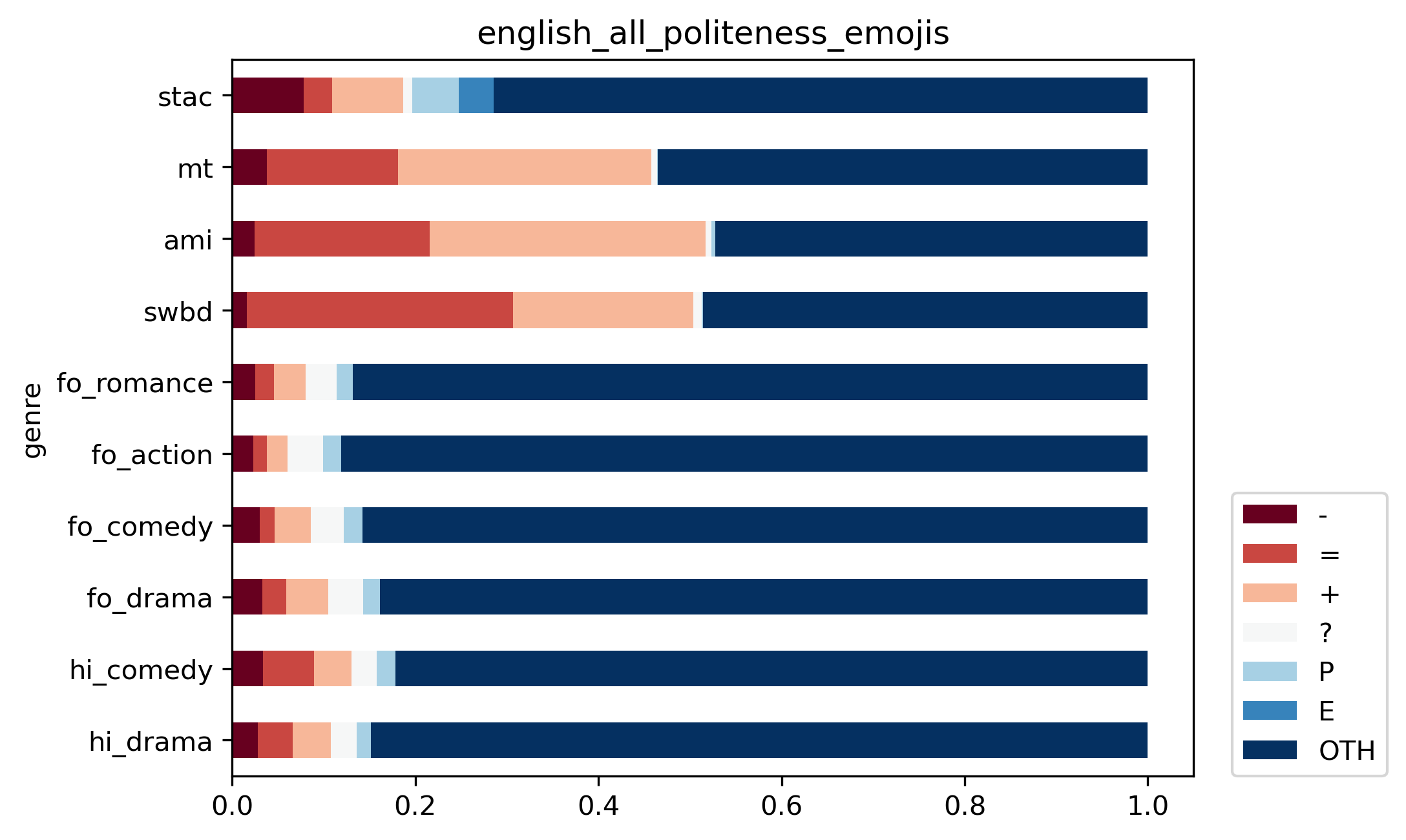}
         \caption{English with politeness ($P$) and Emojis ($E$) keywords}
         \label{fig:en_politeness}
     \end{subfigure}
        \caption{Short utterance distribution including politeness and emojis.}
        \label{fig:politeness}
\end{figure*}

Figures~\ref{fig:lextstats_fr}--\ref{fig:lextstats_ja} present statistics for utterance and feedback types as well as common feedback-related lexical items for different languages. Figure \ref{fig:politeness} shows politeness keywords and emojis in our English and French corpora. \vspace{4mm}


\newpage

\section{Detailed Dialogue Act Tagging Results}
\label{sec:appendix_tagger}

\subsection*{Dialogue Act Grouping}

Table \ref{tbl:swbd_mapped} shows the distribution of instances per mapped dialogue act group in the DAMSL-Switchboard (SWBD) corpus.

\begin{table}[hbt]
\begin{center} 
\begin{tabularx}{\columnwidth}{lrX}  
	\toprule  
	   \textbf{DA group} & \textbf{\# inst.} & \textbf{SWBD labels} \\
	\midrule
	   \bf Forward looking & 109,382 & \atag{sd}, \atag{fx/sv}, \atag{bf}, \atag{na}, \atag{ny\textasciicircum e}, \atag{arp}, \atag{nd}, \atag{no}, \atag{cc}, \atag{co}, \atag{oo}, \atag{ad}, \atag{qr/qy}, \atag{qw}, \atag{qw\textasciicircum d}, \atag{qh}, \atag{qo} \\
	   \bf Backchannel     &  41,017 & \atag{b}, \atag{bk}, \atag{bh}, \atag{bf}, \atag{br} \\
	   \bf Assessment      &  15,727 & \atag{aa}, \atag{fe/ba} \\
	   \bf Yes/no answer   &   4,324 & \atag{ny}, \atag{nn} \\
       \bf Other           &  40,124 & \emph{all other categories} \\
	\midrule
	   \bf Total           & \bf 210,574 &  \\
	\bottomrule
\end{tabularx}
\caption{Instances created from the DAMSL-SWBD corpus with labels mapped to coarse-grained dialogue act groups.} 
    \label{tbl:swbd_mapped} 
\end{center}
\end{table}

\subsection*{Results per Corpus}

Tables~\ref{tbl:pred_conv_detailed} and \ref{tbl:pred_subt_detailed} present the results of our dialogue act tagger per (sub)corpus used. Here, we only make a binary distinction by grouping the feedback-relevant classes \emph{Backchannel} and \emph{Assessment} into a single \emph{Feedback} category. The number of utterances refers to the final version of the data after pre-processing with meta-linguistic information removed. \vspace{3mm}

\begin{table}[h!]
\begin{center}
\begin{tabularx}{\columnwidth}{lXrrr}  
\toprule 
    \bf Lang & \bf Corpus  & \bf \# utt   & \bf \# feedback & \bf \% feedback\\
\midrule
    \bf de & Hamburg MapTask &     4,012 &   1,126 & 28.07 \\
\midrule
    \bf en & AMI     &    83,085 &  20,044 & 24.12 \\
       & Fisher  & 2,117,748 & 421,069 & 19.88 \\
       & HCRC MapTask      &    26,949 &   8,366 & 31.04 \\
       & STAC    &     5,841 &     514 & 8.8 \\
\midrule        
   \bf fr & CID     &    12,326 &   1,754 & 14.23 \\
       & Aix-DVD     &     7,578 &   1,323 & 17.46 \\
       & French MapTask     &     6,046 &   1,226 & 20.28 \\
\midrule
   \bf hu & BUSZI-2 &      30,979 &       6,125 & 19.77 \\
\midrule
   \bf it & CLIPS   &    24,289 &   4,461 & 18.37 \\
\midrule
   \bf ja & Japanese CallHome &     38,701 &      13,432 & 34.71 \\
\midrule
   \bf no & NoTa-Oslo    &    85,506 &  16,861 & 19.72 \\
\midrule
   \bf zh & Chinese CallHome &        17,853 &         2,251 & 12.61 \\
\bottomrule
\end{tabularx}
\caption{\label{tbl:pred_conv_detailed} Number and frequency of communicative feedback phenomena predicted by the BERT-based dialogue act tagger on spontaneous dialogue corpora. Non-English datasets were automatically translated into English with the Google Translate API before inference.} 
\end{center}
\end{table}

{\small
\begin{longtable}{llrrr} 
\toprule 
   \bf Lang & \bf Corpus & \bf \# utt & \bf \# feedback & \bf \% feedback \\
\midrule               
   \bf de & action\_foreign  & 12,760 & 1,703 & 13.35 \\
     & action           & 12,134 & 1,637 & 13.49 \\
     & comedy\_foreign  & 12,627 & 1,849 & 14.64 \\
     & comedy           & 16,152 &  2,369 & 14.67 \\
     & crime\_foreign   & 12,589 & 1,245 & 9.89 \\
     & crime            & 11,817 & 1,581 & 13.38 \\
     & drama\_foreign   & 14,669 & 1,350 & 9.2 \\
     & drama            & 11,460 &  1,452 & 12.67 \\
     & romance\_foreign & 13,499 & 1,500 & 11.11 \\
     & romance          & 11,809 & 1,596 & 13.52  \\
\midrule
   \bf en & action           & 11,094 & 1,437 & 12.95 \\
     & action\_foreign  & 12,908 & 1,448 & 11.22 \\
     & comedy           & 13,948 & 1,665 & 11.94 \\
     & comedy\_foreign  & 13,533 & 1,677 & 12.39 \\
     & crime            & 14,990 & 1,700 & 11.34  \\
     & crime\_foreign   & 13,911 & 1,267 & 9.11 \\
     & drama            & 14,944 & 1,729 & 11.57 \\
     & drama\_foreign   & 10,243 & 1,041 & 10.16 \\
     & romance          & 16,132 & 2,166 & 13.43 \\
     & romance\_foreign & 15,521 & 1,698 & 10.94 \\
\midrule
  \bf fr & action\_foreign  & 11,236 &  1,119 & 9.96 \\
     & action           & 12,406 & 1,453 & 11.71 \\
     & comedy\_foreign  & 17,239 & 1,788 & 10.37 \\
     & comedy           & 13,932 & 1,913 & 13.73 \\
     & crime\_foreign   & 12,159 &   1,017 & 8.36 \\
     & crime            & 10,821 &   1,003 & 9.27 \\
     & drama\_foreign   & 10,002 &   804 & 8.04 \\
     & drama            & 11,094 & 1,313 & 11.84  \\
     & romance\_foreign & 12,043 & 1,360 & 11.29 \\
     & romance          & 13,959 & 1,604 & 11.49 \\
\midrule
   \bf hu & action\_foreign &      12,781 &       1,377 & 10.77 \\
     & comedy\_foreign &      15,031 &       1,998 & 13.29 \\
     & comedy          &      14,692 &       2,462 & 16.76 \\
     & crime\_foreign  &      13,620 &       1,655 & 12.15 \\
     & drama\_foreign  &      13,138 &       1,400 & 10.66 \\
     & drama           &       7,872 &       1,103 & 14.01 \\
     & romance\_foreign &      13,771 &       1,611 & 11.7  \\
\midrule
   \bf it & action\_foreign  & 12,010 &  1,585 & 13.2 \\
     & action           &  7,703 &   826 & 10.72 \\
     & comedy\_foreign  & 15,055 & 2,058 & 13.67 \\
     & comedy           & 15,363 & 1,777 & 11.57 \\
     & crime\_foreign   & 12,454 & 1,320 & 10.6 \\
     & crime            & 13,885 & 1,479 & 10.65 \\
     & drama\_foreign   & 17,444 & 2,289 & 13.12 \\
     & drama            & 12,838 & 1,467 & 11.43 \\
     & romance\_foreign & 14,702 & 1,696 & 11.54 \\
     & romance          & 14,573 & 1,549 & 10.63 \\
\midrule
   \bf ja & action\_foreign &      11,245 &       967 & 8.6 \\
     & action &       3,007 &        443 & 14.73 \\
 & comedy\_foreign &      16,173 &        1,777 & 10.99 \\
      & comedy &      15,675 &       2,555 & 16.3 \\
 & crime\_foreign &      16,296 &       1,311 & 8.04 \\
 & drama\_foreign &      14,201 &        997 & 7.02 \\
       & drama &      11,410 &      1,204 & 10.55 \\
 & romance\_foreign &      14,042 &       1,210 & 8.62 \\
     & romance &       1,780 &        145 & 8.15 \\
\midrule
   \bf no & action\_foreign  & 10,480 &   892 & 8.51 \\
     & action           &  1,855 &   290 & 15.63 \\
     & comedy\_foreign  & 14,406 & 1,834 & 12.73 \\
     & comedy           & 11,957 & 1,199 & 10.03 \\
     & crime\_foreign   & 12,788 & 1,137 & 8.89 \\
     & crime            &  9,863 &   853 & 8.65 \\
     & drama\_foreign   & 12,031 & 1,202 & 9.99 \\
     & drama            &  6,688 &  589 & 8.81 \\
     & romance\_foreign & 12,830 & 1,313 & 10.23 \\
     & romance          &  4,197 &   399 & 9.51 \\
\midrule
   \bf zh & action\_foreign &      11,570 &         967 & 8.36 \\
     & action &       2,722 &        159 & 5.84 \\
     & comedy\_foreign &      14,692 &       1,564 & 10.65 \\
     & comedy &      13,587 &       1,034 & 7.61  \\
     & crime\_foreign &      10,778 &         795 & 7.38 \\
     & crime &      11,182 &       697 & 6.23  \\
     & drama\_foreign &      14,527 &        1,330 & 9.16 \\
     & drama &       9,567 &        743 & 7.77 \\
     & romance\_foreign &      13,362 &       1,079 & 8.08  \\
     & romance &      11,440 &       700 & 6.12 \\
\bottomrule
\caption{Number and frequency of communicative feedback phenomena predicted by the BERT-based dialogue act tagger on  our subtitle corpora. Non-English datasets were automatically translated into English before inference.} 
\label{tbl:pred_subt_detailed} 
\end{longtable}}

\end{document}